\documentclass[preprint,review,12pt]{elsarticle}
\usepackage{amssymb}
\usepackage{amsmath}
\usepackage{graphicx}
\usepackage{url}
\usepackage{multirow}
\usepackage{booktabs}
\usepackage{subfigure}
\usepackage{xcolor}
\usepackage{caption}
\usepackage{epstopdf}
\journal{Patten Recognition}%Neurocomputing}
\begin{document}

\begin{frontmatter}



\title{Multi-Subregion Based Correlation Filter Bank \\
for Robust Face Recognition}


\author{Yan Yan$^{1}$}
\author{Hanzi Wang$^{1}$\corref{cor1}}
\cortext[cor1]{Corresponding author. Tel.: +86-592-2580063; fax: +86-592-2580063.\\
E-mail addresses: yanyan@xmu.edu.cn (Y.~Yan), hanzi.wang@xmu.edu.cn (H.~Wang),\\
dsuter@cs.adelaide.edu.au (D.~Suter).}
\author{David Suter$^{2}$}
\address{$^{1}$School of Information Science and Technology, Xiamen University, China\\
$^{2}$School of Computer Science, The University of Adelaide, Australia}

\begin{abstract}
In this paper, we propose an effective feature extraction algorithm, called Multi-Subregion based Correlation Filter Bank (MS-CFB), for robust face recognition. MS-CFB combines the benefits of global-based and local-based feature extraction algorithms, where multiple correlation filters corresponding to different face subregions are jointly designed to optimize the overall correlation outputs. Furthermore, we reduce the computational complexity of MS-CFB by designing the correlation filter bank in the spatial domain and improve its generalization capability by capitalizing on the unconstrained form during the filter bank design process. MS-CFB not only takes the differences among face subregions into account, but also effectively exploits the discriminative information in face subregions. Experimental results on various public face databases demonstrate that the proposed algorithm provides a better feature representation for classification and achieves higher recognition rates compared with several state-of-the-art algorithms.
\end{abstract}

\end{frontmatter}


\section{Introduction}

In the past few decades, we have witnessed a rapid development of the theories and algorithms of face recognition and its successful applications in access control, video surveillance, law enforcement, human computer interaction, and so on \cite{Zhao2003,Tan2006, Wright2009}. However, face recognition is still a very challenging task due to large face appearance variations caused by occlusions, aging, changes of illumination, facial expression, pose, etc. In particular, in many real-world applications, it often suffers from the small sample size (SSS) problem \cite{Tan2006} since the training samples of each subject are very few, which can severely affect the performance of most face recognition algorithms especially when the dimension of facial feature space is high.
% is one of the most fundamental issues in face recognition

It has been well recognized that effective feature extraction (FE) plays an important role in the success of an face recognition algorithm \cite{Zhao2003,Tan2006,Wright2009,Su2009}. After the FE process, a proper low-dimensional feature vector, with which the class separability is enhanced and the computational complexity of subsequent classifiers is reduced, is generated.
%In the literature, the
FE algorithms can be roughly grouped into two categories \cite{Su2009}: global-based and local-based.  Global-based FE algorithms consider a face region as a whole. The extracted features contain the information embedded in the whole face \cite{Turk1991}. On the other hand,  local-based FE algorithms are based on face subregions (i.e., local facial features, such as eyes, nose, mouth, and chin \cite{Su2009,Kumar2011,Timo2006})
 %or the face region is divided into blocks of the same size \cite{})
 and encode the detailed characteristics within each face subregion.
% Just as the human perception system, the hybrid-based methods use both local and the whole face region to extract a feature. The hybrid methods could potentially offer % the best of the global and local methods.

Traditional local-based FE algorithms usually combine the outputs from different face subregions by adopting a fusion strategy (e.g., the majority voting \cite{Zhu2012}, the weighted sum \cite{Su2009,Kanade2003,Bonnen2012}, or the concatenation of original$/$low-dimensional features \cite{Kim2005,Heisele2007,Li2010}). Note that the above-mentioned algorithms consider the local FE step and the combination of different subregions as two independent processes. Although many successful local-based FE algorithms have been proposed, it remains an open issue that how to combine these two processes as a whole.

%Unfortunately, this method may not work well since the features of various modalities may not be compatible [1].
%Recently, multimodal biometrics fusion techniques have attracted much attention in the belief that the supplementary information between different modalities can improve the recognition performance. Many works have concentrated on this area [1-4]. In general, these methods can be classified into three categories: fusion at the feature level, fusion at the match level and fusion at the decision level [1]. In this paper, we mainly focus on fusion at the feature level which is believed to be very promising since feature sets can provide more information about the input biometrics than other levels [1,2].
%On the other hand, subspace learning methods, e.g. Principal Component Analysis (PCA) [5], Linear Discriminant Analysis (LDA) [6], Locality Preserving Projections (LPP) [7] and Class-dependence Feature Analysis (CFA) [8], which select low-dimensional features to represent raw data, have been widely used in the biometric researches. Unlike traditional subspace learning methods [5-7], the projection axis obtained by CFA which is based on the design of correlation filter (CF) technique tries to discriminate one specific class from all other classes. According to different criterions, different correlation filters can be designed.
%Subspace learning methods which perform at the feature level for multimodal biometrics is usually implemented by concatenating two or more original or low-dimensional features to form a long vector [3].

In this paper, we propose an effective feature extraction algorithm, called Multi-Subregion based Correlation Filter Bank (MS-CFB), for robust face recognition. A new type of  filter bank, i.e., Correlation Filter Bank (CFB), is employed in MS-CFB.  We formulate the filter bank design as a minimization problem of the generalized Rayleigh quotient \cite{Golub1996}, which has a closed-form solution. The advantages of this development are the reduction in the computational complexity and the simplification in the decision process, since we can obtain multiple correlation filters corresponding to different face subregions simultaneously.
%the filter bank has a closed-form solution.
%In CFB, the unconstrained correlation filter trained for a specific face region is designed by optimizing the overall origin correlation outputs. Hence, CFB takes full advantage of the information in different subregions.
%Furthermore, CFB is significantly effective in dealing with the occlusion due to the summation mechanism used during the feature vector extraction.

Compared with traditional algorithms, the proposed MS-CFB algorithm has the following characteristics:
\begin{itemize}
        \item
        MS-CFB makes use of local facial features to perform global FE. Therefore, MS-CFB exploits the benefits of both local face subregions and the whole face for extracting features, which incorporates the advantages of both global-based and local-based FE algorithms.
        %In MS-CFB, the correlation filters designed for different face subregions are jointly trained by optimizing the overall %correlation outputs. Therefore, not only are the differences between face subregions taken into account, but also the %useful information in various face subregions is fully exploited.
        \item
        Traditional local-based FE algorithms consider the local FE step and the combination of different face subregions as two independent processes. In contrast, MS-CFB tries to unify these two processes in an integrated framework. The local FE step of MS-CFB aims to optimize the overall correlation outputs from all face subregions. Such strategy enhances the effectiveness of local feature extraction.
%        On one hand, all the correlation filters trained for different facial subregions are simultaneously designed by optimizing the overall origin correlation outputs. On the other hand,
        %The outputs from the different face subregions are combined together to form the global feature vector.
        %While the local feature extraction and the combination of different subregions are considered as two separate problems in the traditional local-based face recognition methods, our proposed MS-CFB tries to solve two problems simultaneously in the CFA framework. The feature extraction in different facial subregions aim
%        aims to
%        combines the two processes as whole
%
%        aim to combine the outputs of different facial subregions by using a simple strategy, our proposed MS-CFB
%        Compared with traditional fusion methods at the feature level which concatenate features to form a long vector, the dimensionality of the features obtained by the novel method is equal to the number of classes in the training set no matter how many regions are used.
        \item
        %We derive the CFB based on the spatial domain. The importance of this development is that it allows us
        While conventional correlation filters \cite{Kumar1999} rely on the frequency domain representations, the design process of a CFB is based on the spatial domain representations, which effectively reduces the computational complexity during the filter bank design process (this is because the Fourier transforms used in traditional algorithms are not required). Moreover, compared with commonly used constrained correlation filters in face recognition (such as OTF \cite{Kumar1999}), a CFB is designed by capitalizing on the unconstrained form to improve its generalization capability.
\end{itemize}

The remainder of this paper is organized as follows.~Related work is discussed in Section 2.~A detailed description of the proposed MS-CFB algorithm is presented in Section 3.~In Section 4, the experimental results on various public face databases are given.~Finally, the concluding remarks and future work are provided in Section 5.%and suggestions for future work
\section{Related Work}
In this section, we begin with reviewing some widely used FE algorithms including popular global-based and local-based FE algorithms in Section 2.1. Some traditional and recently developed correlation filters are described in Section 2.2. The motivation of this work is given in Section 2.3.
\subsection{Global-based and Local-based FE Algorithms}
\begin{figure*}[tbh!]
         \centering
         \subfigure[LDA]{
            \includegraphics[width=6.5cm,height=5cm]{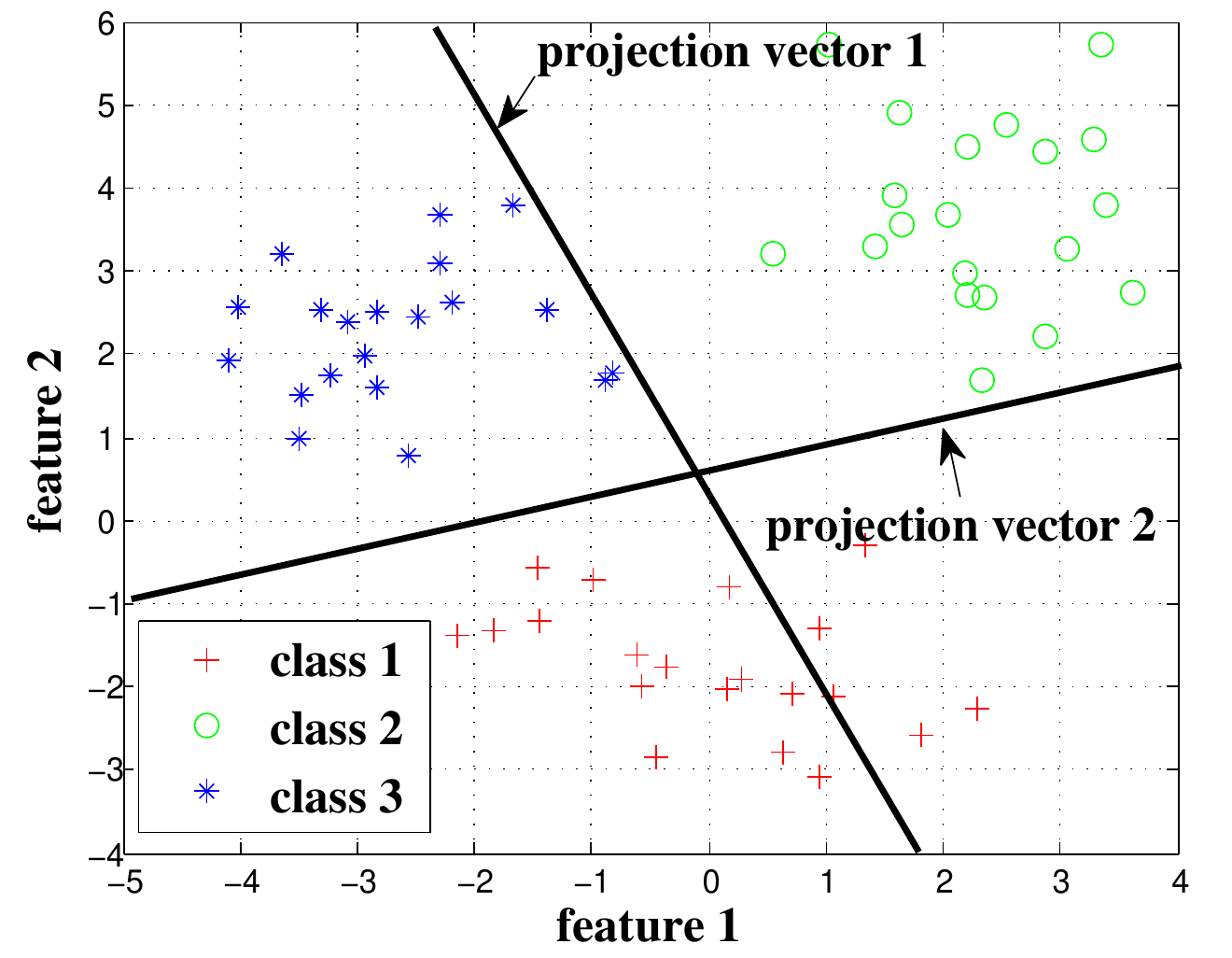}%
            }
          \subfigure[CFA]{
            \includegraphics[width=6.6cm,height=5cm]{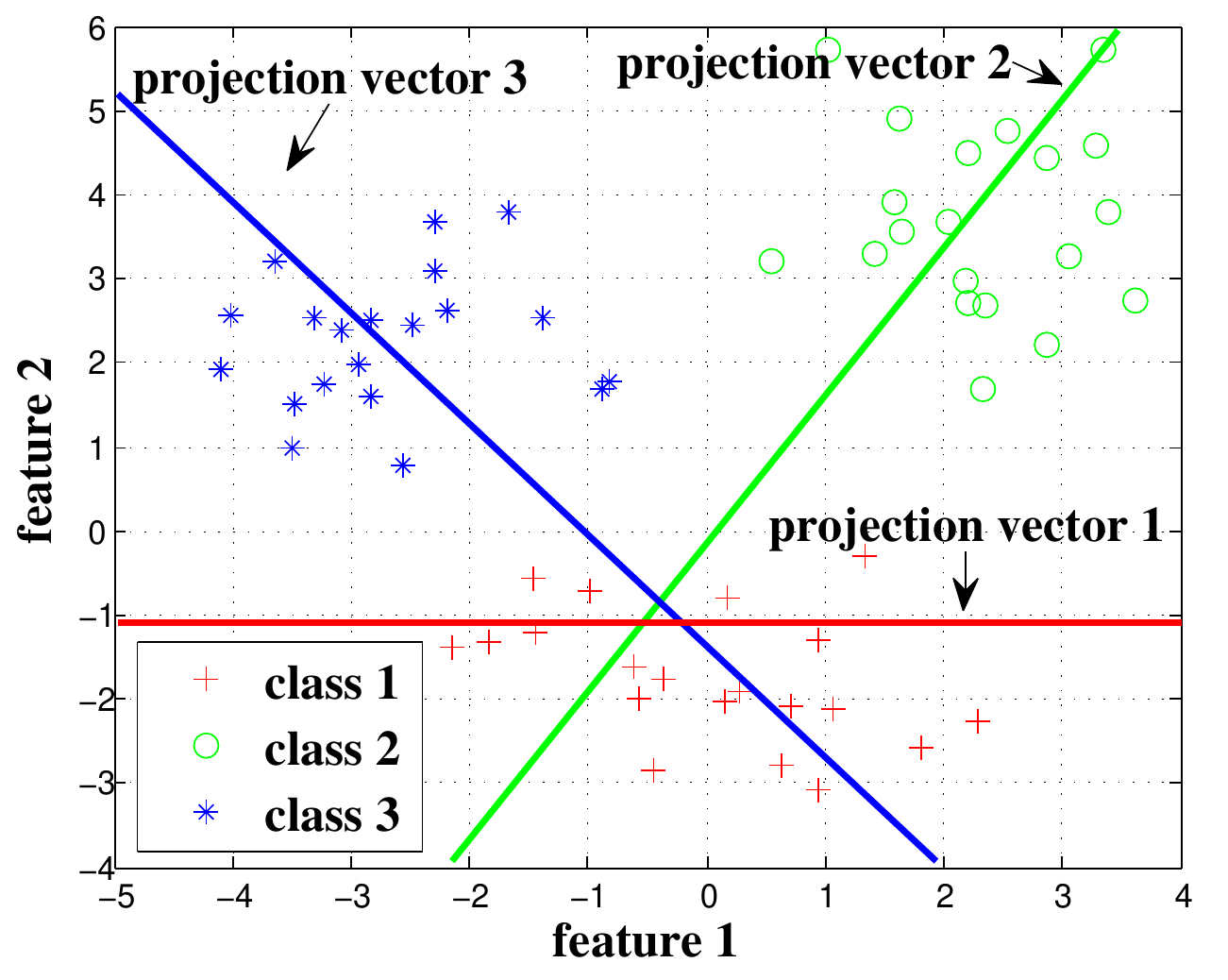}%
            }
         \caption{A comparison of the projection vectors obtained by (a) LDA and (b) CFA for a three-class problem. Each projection vector obtained by LDA discriminates all three classes while that obtained by CFA discriminates one class from the other two classes. Note that LDA obtains only two projection vectors.
                 }
         \label{FIG:CFALDA}%
         \end{figure*}

A large number of global-based FE algorithms have been developed so far. One of the most successful algorithms for face recognition is  appearance-based algorithms, where a face is represented as a vector (e.g., it can be obtained by concatenating each row/column of a face image) \cite{Turk1991,Belhumeur1997,He2005} or a tensor \cite{Tao2007,Yan2007}. In practice, however, a high-dimensional vector or a tensor are too large to allow fast and robust face recognition.~A common way to solve this problem is to use dimensionality reduction algorithms, such as Principal Component Analysis (PCA) \cite{Turk1991}, Linear Discriminant Analysis (LDA) \cite{Belhumeur1997,Tao2007}, or Class-dependence Feature Analysis (CFA) \cite{Kumar2006,Yan2008}. Each projection vector in the projection matrix obtained by PCA (or LDA) tries to represent (or discriminate) all classes in the new feature space. On the other hand, each projection vector obtained by CFA, which is based on the design of the correlation filters, discriminates one class from all the other classes. Fig.~\ref{FIG:CFALDA} shows a comparison of the projection vectors obtained by LDA and CFA for a three-class problem.
%Recently, the sparse representation based classification (SRC) has been successfully used in FR. SRC takes into the account %the fact that the high-dimensional face images can be sparsely represented by the representative samples. And the sparse %coding coefficients

%-----------------------------------------------------------
%A normalization step is carried out prior to the concatenation of the LDA coefficients.
Global-based FE algorithms, however, do not consider the diversity of local facial structures which can be useful for classification. Recently, local-based FE algorithms have received much attention due to the fact that local facial features (such as eyes and mouth) are more robust to variations of illumination, facial expression, and pose. In \cite{Penev1996}, the Local Feature Analysis (LFA) algorithm was introduced to encode the local topographic representations of a face image, where kernels of local spatial support are used to extract information from local face subregions.
Kim et al.~\cite{Kim2005} presented a component-based LDA FE algorithm for image retrieval. Each face subregion is firstly represented as the LDA coefficients in the Fisher subspace. Then, a feature vector is formulated by concatenating all of the coefficients. Finally, a holistic LDA \cite{Belhumeur1997}, which reduces the dimension of the combined feature vectors, is employed to obtain a compact representation. Li et al.~\cite{Li2010} proposed a Block-based Bag Of Words (BBOW) algorithm for robust face recognition. Dense SIFT features \cite{Lowe2004} are calculated and quantized into different codewords for each face subregion.~Then, histograms of each face subregion are concatenated to obtain a feature vector. Finally, linear SVM classifiers are employed to perform classification. Su et al.~\cite{Su2009} proposed a novel face recognition algorithm which employs both global and local classifiers. The global feature vector is extracted from a whole face image by using the low frequency Fourier coefficients, while the local feature vector is constructed based on LDA. The final classifier is formed by combining (i.e., using the linear weighted sum) a global feature based classifier and a local feature based classifier.
  %(formed by weighting multiple LDAs).
  Zhu et al.~\cite{Zhu2012} proposed a Patch-based Collaborative Representation based Classification (PCRC) algorithm for face recognition. The majority voting of the classification outputs from all face subregions is employed to make a final decision. Furthermore, in order to make PCRC less sensitive to the size of face subregions, a multi-scale scheme is used by integrating the complementary information obtained at different scales.

 We should point out that, in this paper, we focus on the FE technique, mainly referred to dimensionality reduction \cite{Yan2007}, which aims to find a mapping from a high-dimensional image space onto a desired low-dimensional face subspace in a global or local manner.
 %an FE algorithm mainly refers to

 %As a matter of fact, original face representations can also be global (e.g., pixel intensity \cite{Turk1991} and image gradient orientations \cite{Tzimiropoulos2012}) or %local (e.g., SIFT \cite{Lowe2004}, LBP \cite{Timo2006}, and Gabor \cite{Liu2004}).
 \subsection{Correlation Filters}
Since the pioneering work by VanderLugt \cite{VanderLugt1964}, correlation filters have been widely used in signal processing and pattern recognition for decades. One of the most simple correlation filters is the Matched Filter (MF) \cite{VanderLugt1964,Choi1997}, which uses the complex conjugate of a reference sample. An MF is optimal only when an input sample and the reference sample are identical except that they are with different white noises.
However, for practical applications, an input sample suffers from different variations, such as rotations and illumination changes, and thus an MF does not perform well.~Therefore, the composite correlation filters \cite{Kumar2006} were developed instead of a single correlation filter. For instance,
Hester et al.~\cite{Hester1982} proposed the concept of the Synthetic Discriminant Function (SDF) filter, which is the weighted sum of MFs. An SDF filter produces high correlation peaks for authentic samples but it does not consider impostor samples. A Minimum Average Correlation Energy (MACE) filter \cite{Mahalanobis1987} was proposed to minimize the average energy of a correlation plane for all samples while constraining the correlation outputs for authentic samples. However, an MACE filter emphasizes high frequency parts of samples, which makes it susceptible to noise. An Optimal Tradeoff Filter (OTF) \cite{Refregier1990} was designed by combining a Minimum Variance Synthetic Discriminant Function (MVSDF) filter \cite{Kumar1986} (focusing on the low frequency parts of samples) and an MACE filter.
%Recently, Zhu et al. \cite{Zhu2007} developed the Feature Correlation Filter (FCF) by extending the concept of correlation filter to feature spaces.
Yan et al.~\cite{Yan2008} proposed an Optimal Extra-class Output Tradeoff Filter (OEOTF) to emphasize the outputs for extra-class samples.
\subsection{Motivation}
%needs to be re-written with a more comprehensive literature survey and then more clearly delineate what is new here and why
%it should be a better system.
Recent studies \cite{Zhao2003,Su2009} have suggested that a hybrid-based FE algorithm, which makes use of both global-based and local-based FE algorithms, could potentially offer the best of the two types of algorithms. Hence, in this paper we combine global-based and local-based FE algorithms in a principled way. Here, instead of extracting local facial features separately and then combining them by using the weighted sum or the majority voting, the proposed algorithm directly extracts a global feature vector based on the combination of local features. Meanwhile, the local FE steps for different face subregions are jointly performed so that
the overall correlation outputs from all face subregions satisfy the design criterion.

%the local feature extraction step for different subregions is optimized so that generates the optimal results.

On the other hand, to adapt to the correlation filter which is specifically designed for the face recognition task, instead of optimizing the whole correlation plane, we propose to optimize the origin peaks in the correlation plane. This improvement is motivated by the fact that the proposed feature extraction framework mainly considers the information of the origin peaks. One merit of working on the origin peaks is that traditional Fourier transforms are not required (based on the generalized Parseval's theorem \cite{Oppenheim2009}), which improves the computational efficiency during the design process.

\section{Multi-Subregion Based Correlation Filter Bank (MS-CFB)}
In this section, an overview of the proposed MS-CFB algorithm for face recognition is introduced in Section 3.1. The detailed design process of a CFB and feature extraction based on CFBs are described in Sections 3.2 and 3.3, respectively. Classification rule is presented in Section 3.4. The complete algorithm is given in Section 3.5. We discuss the proposed algorithm in Section 3.6.

Before formally presenting the proposed algorithm, we begin by introducing the notations used in this paper. Light case symbols represent the spatial domain while bold case ones refer to the frequency domain.
\subsection{Overview of the MS-CFB Algorithm for Face Recognition }
\begin{figure*}[tbh!]
         \centering
         \scalebox{0.88}{
            \includegraphics[width=10cm,height=10cm]{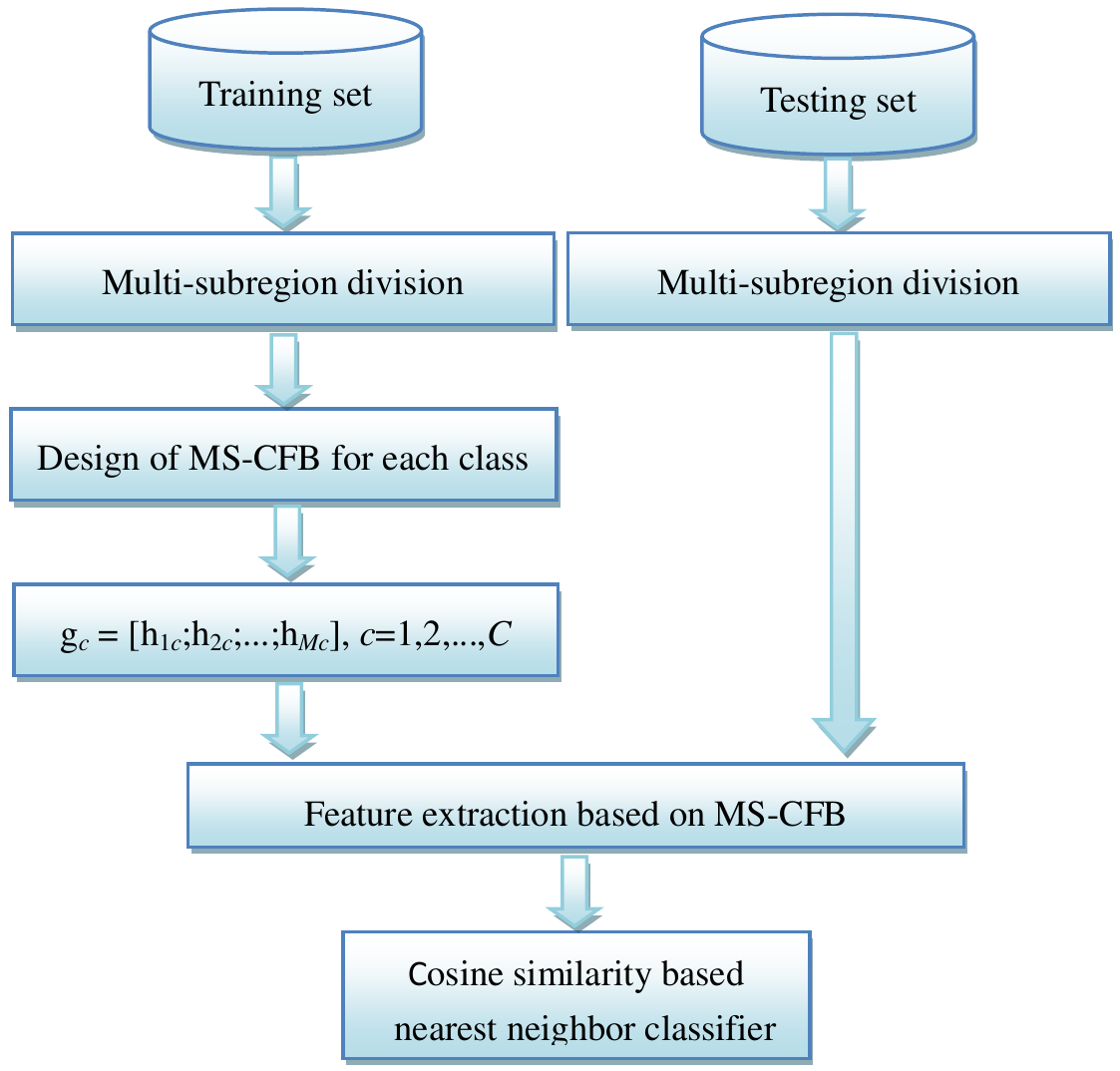}
           }
         \caption{An overview of the MS-CFB algorithm for face recognition.
                 }
         \label{FIG:System}%
        \end{figure*}

An overview of the proposed MS-CFB algorithm for face recognition is shown in Fig.~\ref{FIG:System}.

Inspired by CFA \cite{Kumar2006,Yan2008}, the proposed algorithm tries to distinguish one class from all the other classes for each projection vector. During the training stage, for each face image in the training set, it is firstly divided into multiple blocks of the same size (corresponding to different face subregions).
Each face subregion is represented as a high-dimensional vector by concatenating the pixel values in the subregion (other face feature representations, such as SIFT \cite{Lowe2004} and Gabor \cite{Liu2004}, can also be used). Secondly, a set of Correlation Filter Banks (CFBs) is designed for all classes (see Section 3.2) and then used to perform feature extraction (see Section 3.3). More specifically, a class-specific CFB is designed for each class in the training set to discriminate that class from all the other classes, and thus a set of class-specific CFBs is obtained for all classes and employed to extract features. During the test stage, for a face image in the test set, after the multi-subregion division procedure, a feature vector is extracted based on CFBs. Finally, a nearest neighbor classifier is employed for classification.

\subsection{Design Process of a CFB}
%The set of correlation filters in MS-CFB for class $c$
%is represented as $\{h_{1c},h_{2c},\cdots,h_{Mc}\}$, where $M$ is the number of face subregions.
        \begin{figure}%[tbh!]
         \centering
         \scalebox{0.85}{
            \includegraphics[width=16.5cm,height=10cm]{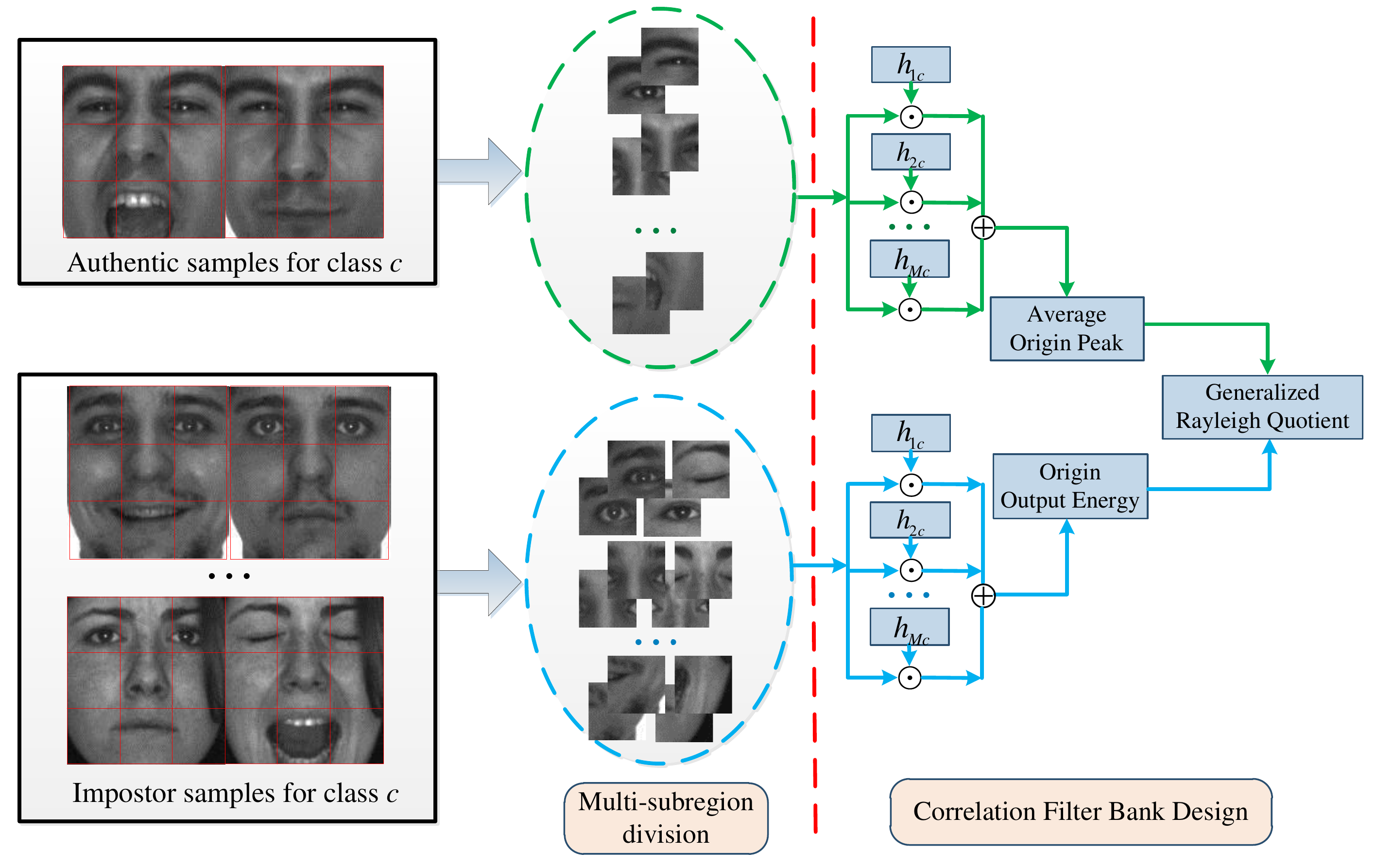}
            }
         \caption{The design process of a CFB. `$\odot$' represents the inner product.
                 }
         \label{FIG:CFB}%
        \end{figure}
%-----------------------------------------------------------

Assume that there are $N$ training images and $C$ classes in the training set. We aim to design a CFB for class $c$ ($c={1,2,\cdots,C}$). The design process of a CFB for class $c$ is shown in Fig.~\ref{FIG:CFB}.

First, we define the overall correlation output ($O[n]$) of a CFB as:
\begin{equation}
\label{EQ:1}
O[n] = \sum_{m=1}^{M}(x_{m}\otimes h_{m,c})[n],
\end{equation}
where $x_{m}$ is the raw feature vector of the $m$-th face subregion; $h_{m,c}$ is a correlation filter corresponding to the $m$-th face subregion for class $c$; $M$ is the number of face subregions in a face image; `$\otimes$' stands for the correlation operator.
%$x_{m}$ is the raw feature of the $m$-th face subregion, and $h_{mc}$ is the correlation filter corresponding to the $m$-th face subregion for class $c$.

According to the Fourier transform theory \cite{Oppenheim2009}, the above equation can be re-written in the frequency domain, that is:
\begin{equation}
\label{EQ:2}
O[n] = \sum_{m=1}^{M}\sum_{k=0}^{D-1}(\textbf{X}_{m}[k])^{*}\textbf{H}_{m,c}[k]e^{\frac{j2\pi kn}{P}},
\end{equation}
where $\textbf{X}_{m}[k]$ and $\textbf{H}_{m,c}[k]$ are the Fourier transforms of $x_{m}$ and $h_{m,c}$, respectively; `$*$' denotes the conjugate operator; $n$ and $k$ represent the indexes in the spatial domain and frequency domain, respectively; $D$ is the dimension of the raw local facial feature vector. Note that the point $O[0]$, which is equal to  the sum of the inner products between the inputs and the correlation filters, is usually referred to the overall origin correlation output or the overall origin peak.

In the CFB, all of the correlation filters are jointly designed %to provide the best performances
so that the outputs for authentic training samples (refer to the training samples in class $c$) and the ones for impostor training samples (refer to the training samples that are excluded from class $c$) are well separated.
To achieve this goal, we emphasize the outputs for authentic training samples while at the same time, suppressing the outputs for impostor training samples.
Formally,  the design criterion of a CFB is to minimize the overall origin output energy for impostor training samples and simultaneously maximize the average overall origin peak for authentic training samples for the class of interest.
%In this paper, MS-CFB is designed by optimizing the overall origin correlation outputs. Particularly,

According to Eq.~(\ref{EQ:2}), the overall origin output energy ($\mathbb{E}_{I}$) for impostor training samples of class $c$ can be derived as:
\begin{eqnarray}
\label{EQ:3}%
\mathbb{E}_{I}
&=& \frac{1}{N_{c}^{I}}\sum_{i=1}^{N_{c}^{I}}|O^{I}_{i,c}[0]|^{2} \nonumber \\
&=& \frac{1}{N_{c}^{I}}\sum_{i=1}^{N_{c}^{I}}|\sum_{m=1}^{M}\sum_{k=0}^{D-1}(\textbf{X}_{mi,c}^{I}[k])^{*}\textbf{H}_{m,c}[k]|^{2},
\end{eqnarray}
where $O^{I}_{i,c}[0]$ represents the overall origin correlation output corresponding to the $i$-th impostor training sample of class $c$;
$\textbf{X}_{mi,c}^{I}[k]$ is the Fourier transform of $x_{mi,c}^{I}$ (see the definition below); $N_{c}^{I}$ is the number of impostor training samples of class $c$.

Based on the generalized Parseval's theorem \cite{Oppenheim2009} (which shows that the correlation of two functions is equal to the product of the individual Fourier transforms of the functions, where one of them is complex conjugated), in Eq.~(\ref{EQ:3}) we can replace the representations of features in the frequency domain with those in the spatial domain. Therefore, the right side of Eq.~(\ref{EQ:3}) is equivalent to the following equation:
\begin{eqnarray}
\label{EQ:4}%
\frac{1}{N_{c}^{I}}\sum_{i=1}^{N_{c}^{I}}|D\sum_{m=1}^{M}\sum_{n=0}^{D-1}x_{mi,c}^{I}[n]h_{m,c}[n]|^{2}
&=& \frac{D^2}{N_{c}^{I}}\sum_{i=1}^{N_{c}^{I}}|\sum_{m=1}^{M}h_{m,c}^{\rm{T}}x_{mi,c}^{I}|^{2} \nonumber \\
&= &\frac{D^2}{N_{c}^{I}}\sum_{i=1}^{N_{c}^{I}}g_{c}^{\rm{T}}(X_{i,c}^{I})(X_{i,c}^{I})^{\rm{T}}g_{c} \nonumber \\
&=& g_{c}^{\rm{T}}\Sigma_{c}g_{c},
\end{eqnarray}
where $x_{mi,c}^{I} = (x_{mi,c}^{I}[0],x_{mi,c}^{I}[1],\cdots,x_{mi,c}^{I}[D-1])^{\rm{T}}$ is the raw feature vector corresponding to the $m$-th face subregion of the $i$-th impostor training sample of class $c$;
$h_{m,c} = (h_{m,c}[0],h_{m,c}[1],\cdots,h_{m,c}[D-1])^{\rm{T}}$ is the corresponding correlation filter; $X_{i,c}^{I} = (x_{1i,c}^{I};x_{2i,c}^{I};\cdots;x_{Mi,c}^{I})$$\in R^{MD\times 1}$ is a column vector, which contains $M$ different face subregions of the $i$-th impostor training sample; $g_{c} = (h_{1,c};h_{2,c};\cdots;h_{M,c})$$\in R^{MD\times 1}$ is composed of $M$ correlation filters corresponding to $M$ face subregions, and
\begin{eqnarray}
\label{EQ:RC}%
\Sigma_{c} = \frac{D^2}{N_{c}^{I}}\sum_{i=1}^{N_{c}^{I}}(X_{i,c}^{I})(X_{i,c}^{I})^{\rm{T}},
\end{eqnarray}
where $\Sigma_{c}$ is the covariance matrix which effectively encodes the relationships among $M$ different face subregions.

The average overall origin peak ($\mathbb{P}_{A}$) for authentic training samples of class $c$ can be expressed as:
\begin{eqnarray}
\label{EQ:5}
\mathbb{P}_{A} &=& \frac{1}{N_{c}}\sum_{j=1}^{N_{c}}O^{A}_{j,c}[0] \nonumber \\
&=& \frac{D}{N_{c}}\sum_{j=1}^{N_{c}}\sum_{m=1}^{M}\sum_{n=0}^{D-1}x_{mj,c}^{A}[n]h_{m,c}[n],
\end{eqnarray}
where $O^{A}_{j,c}[0]$ represents the overall origin correlation output corresponding to the $j$-th authentic training sample of class $c$; $N_{c}$ is the number of authentic training samples of class $c$.

Using the vector representation, the right side item of Eq.~(6) can be converted as:
\begin{eqnarray}
\label{EQ:6}
\frac{D}{N_{c}}\sum_{j=1}^{N_{c}}\sum_{m=1}^{M}\sum_{n=0}^{D-1}x_{mj,c}^{A}[n]h_{m,c}[n]&=& \frac{D}{N_{c}}\sum_{j=1}^{N_{c}}\sum_{m=1}^{M}h_{m,c}^{\rm{T}}x_{mj,c}^{A} \nonumber \\
&=&  \frac{D}{N_{c}}\sum_{j=1}^{N_{c}}(X_{j,c}^{A})^{\rm{T}}g_{c}  \nonumber \\
&=&  m_{c}^{\rm{T}}g_{c},
\end{eqnarray}
where $x_{mj,c}^{A} = (x_{mj,c}^{A}[0],x_{mj,c}^{A}[1],\cdots,x_{mj,c}^{A}[D-1])^{\rm{T}}$ is the raw feature vector  corresponding to the $m$-th face subregion of the $j$-th authentic training sample of class $c$; $X_{j,c}^{A} = (x_{1j,c}^{A};x_{2j,c}^{A};$ $\cdots;x_{Mj,c}^{A}) \in R^{MD\times 1}$ is a column vector, which contains $M$ different face subregions of the $j$-th authentic training sample, and
\begin{eqnarray}
\label{EQ:mean}
m_{c} = \frac{D}{N_{c}}\sum_{j=1}^{N_{c}}X_{j,c}^{A},
\end{eqnarray}
where $m_{c}$ is the mean of all authentic training samples of class $c$.

Therefore, in order to maximize the average overall origin peak for authentic training samples while minimizing the overall origin output energy for impostor training samples, we employ the quotient form
by combining Eqs.~(4) and (7), that is,
\begin{equation}
\label{EQ:7}
J(g_{c}) = \frac{\mathbb{P}_{A}^{2}}{\mathbb{E}_{I}} = \frac{|m_{c}^{\rm{T}}g_{c}|^{2}}{g_{c}^{\rm{T}}\Sigma_{c}g_{c}}.
\end{equation}%\text{arg} \max_{g_{c}}

As we can see, $J(g_{c})$ is the generalized Rayleigh quotient \cite{Golub1996} which reaches its maximal value when $\Sigma_{c}$ is a non-singular matrix.
Unfortunately, recalling that $\Sigma_{c} = \frac{D^2}{N_{c}^{I}}\sum_{i=1}^{N_{c}^{I}}(X_{i,c}^{I})(X_{i,c}^{I})^{\rm{T}}$, where $X_{i,c}^{I}\in R^{MD\times 1}$, it is easy to derive that $\Sigma_{c}\in R^{MD\times MD}$ is a singular matrix since $rank(\Sigma_{c}) \leq N_{c}^{I}$ (by using the properties of the rank) and $N_{c}^{I} \ll MD$ (i.e., the SSS problem).
Therefore, to resolve the singularity problem of $\Sigma_{c}$, we add a regularized term to Eq.~(9). As a result, the optimization criterion becomes
\begin{equation}
\label{EQ:8}
g_{c} = \text{arg} \max_{g_{c}}\frac{|m_{c}^{\rm{T}}g_{c}|^{2}}{g_{c}^{\rm{T}}\hat{\Sigma}_{c}g_{c}}.
\end{equation}%\text{arg} \max_{g_{c}}
Here $\hat{\Sigma}_c = (1-\alpha)\Sigma_{c}+\alpha\mathbb{I}$, where
$\alpha$ $(\in [0,1])$ is the regularized parameter and $\mathbb{I} \in R^{MD\times MD}$ is an identity matrix.

Based on some matrix operations \cite{Golub1996}, the solution of Eq.~(10) is
\begin{equation}
\label{EQ:9}
g_{c} = \hat{\Sigma}_c^{-1}m_{c}.
\end{equation}

Once $g_{c}$ is computed, all of the correlation filters $h_{m,c}$  $(m=1,2,\cdots,M)$  can be obtained simultaneously for class $c$.
        \begin{figure}%[tbh!]
         \centering
         %\scalebox{0.5}{
            \includegraphics[width=12cm,height=8cm]{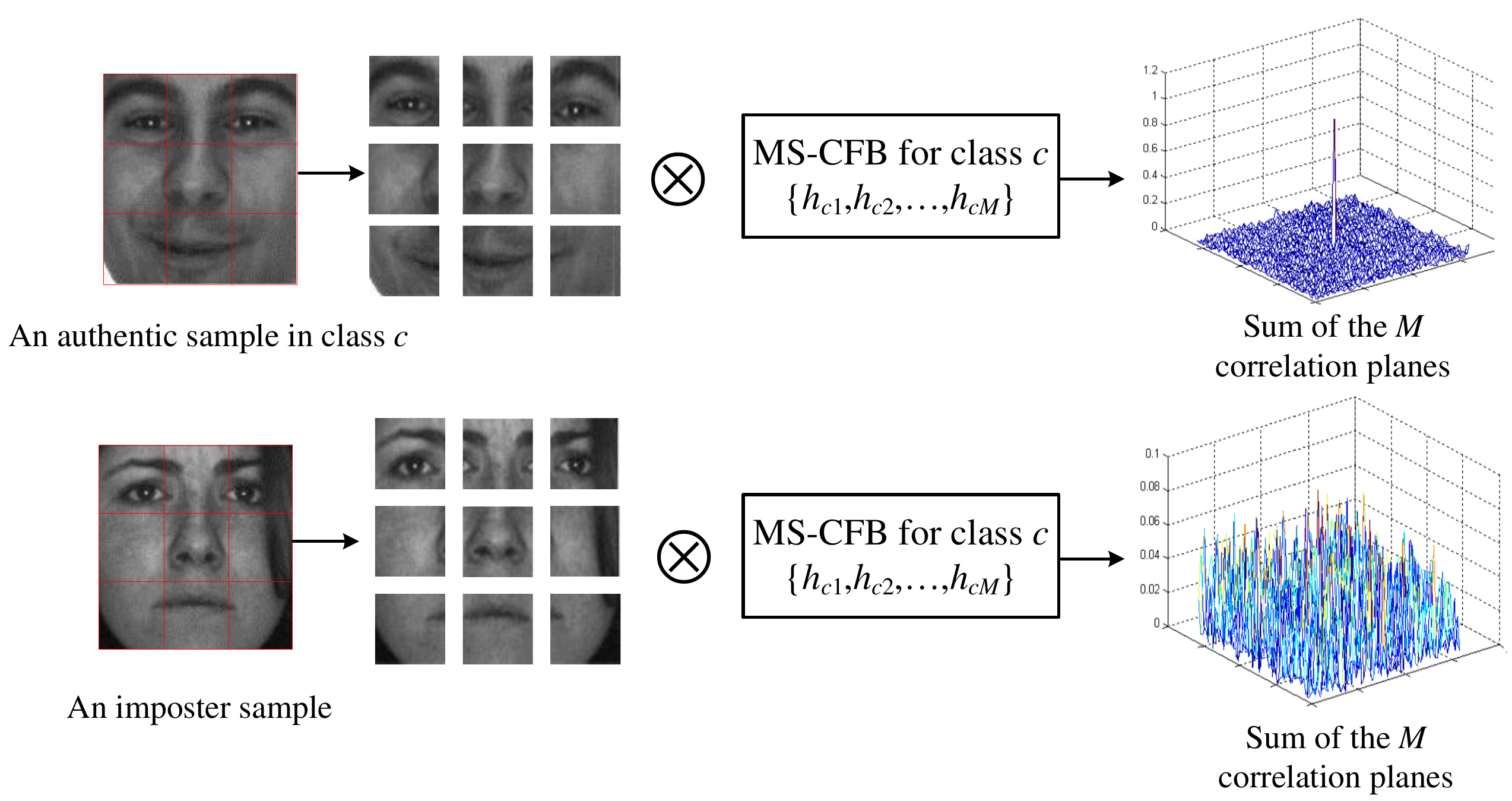}
          %  }
         \caption{The correlation outputs of a CFB for an authentic test sample and an impostor one.
                 }
         \label{FIG:Discussion}%
        \end{figure}
%-----------------------------------------------------------
In this paper, $g_{c}$ is termed as the Correlation Filter Bank (CFB), since it consists of multiple correlation filters corresponding to different face subregions. Fig.~\ref{FIG:Discussion} illustrates the typical correlation outputs of a CFB for an authentic test sample and an impostor one. As shown in Fig.~\ref{FIG:Discussion}, for an authentic test sample, the CFB can produce a sharp peak, while the
correlation output has no discernible peak for an impostor test sample.

The time complexity to design a CFB is $O(N_{c}^{I}(MD)^{2} + (MD)^{3} + MD)$, where $M$ and $D$ are the number of face subregions and the dimension of local facial feature space, respectively.
$N_{c}^{I}$ is the number of impostor training samples of class $c$. The time cost mainly comprises three parts:
$O(N_{c}^{I}(MD)^{2})$ is used to compute $\hat{\Sigma}_c$; $ O((MD)^{3})$ is used to calculate the matrix inversion of $\hat{\Sigma}_c$; and
$O(MD)$ is used to construct the final $g_{c}$. Therefore, the non-diagonal matrix inversion of $\hat{\Sigma}_c$ consumes the majority time during the design process of a CFB.
%\subsection{Similarity Measure}
% Why not Maximum Value
%
%However, the direct use of Eq. (14) for classification has two drawbacks: 1)  it discards the information in other components of the feature vector which is beneficial for classification; and 2) the above-mentioned classification rule is not valid when it is used to classify unseen subjects. Different from conventional pattern recognition tasks, the subjects in the test set may not exist in the training set.
%In such scenarios, each component in the  feature vector extracted by our algorithm represents the identity similarity between the training subject and the test one. Thus, all components are required to measure the similarity. As a result, to overcome the above two problems, we use the cosine similarity measure for the NN classifier.
\subsection{Feature Extraction Based on CFBs}
%where we extract a feature for each block and construct a global feature %vector that represents both the statistics of the local facial subregions %and the spatial locations. Then,
After obtaining a set of CFBs (a CFB is designed by optimizing Eq.~(\ref{EQ:8}) for one class) during the training stage, we can perform feature extraction for both training set and test set. A face image correlated with all CFBs generates a features vector to represent the image.

\begin{figure}%[tbh!]
         \centering
         \scalebox{0.85}{
            \includegraphics[width=17.5cm,height=9cm]{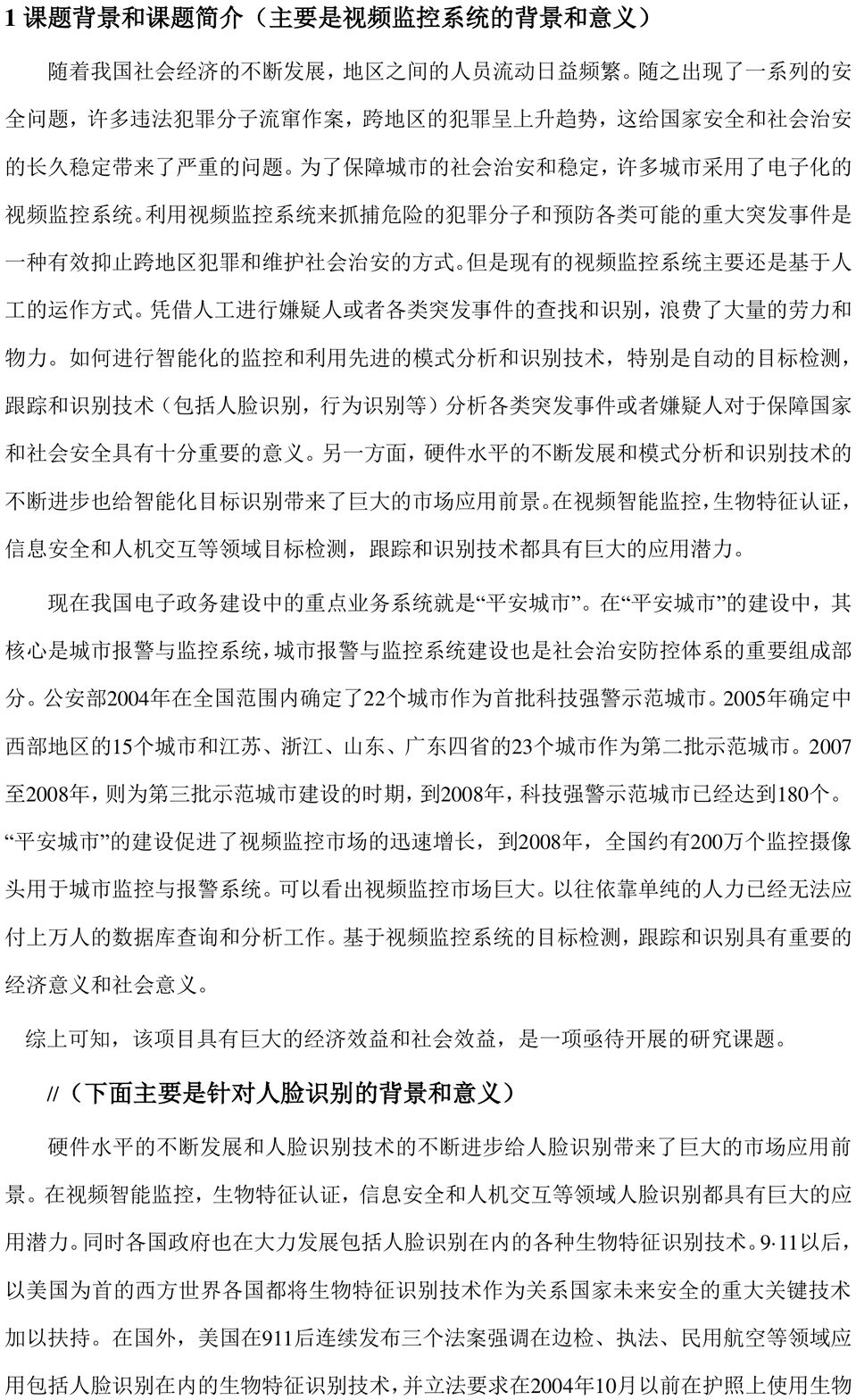}
          }
         \caption{The proposed framework of feature extraction based on CFBs. `$\otimes$' and `$\oplus$' represent the correlation operator and summation operator, respectively.
                 }
         \label{FIG:Framework}%
        \end{figure}

The proposed framework of feature extraction based on CFBs is illustrated in Fig.~\ref{FIG:Framework}.
%Suppose that we have obtained a set of CFBs during the training stage,
 %both and spatial locations of the subregions
According to Fig.~\ref{FIG:Framework}, the sum of the correlation outputs is first computed for each CFB. A global feature vector, which exploits the statistics of local face subregions, is then constructed based on the origin correlation outputs of all CFBs. To be specific, after the multi-subregion division procedure, a raw feature vector is first extracted for each face subregion. Next, the correlations between the correlation filters in the CFB and the corresponding raw feature vectors are calculated and then summed for each face class. Finally, a global feature vector is obtained, whose components respectively represent the overall origin correlation outputs of all CFBs in the summed correlation output plane.
In fact, the overall origin correlation output can also be derived by cumulating the inner products between the local features and a CFB. Mathematically, after obtaining a set of CFBs for all classes,
each component in a global feature vector $f = (f[1],f[2],\cdots,f[C])^{\rm{T}}$ can be obtained by
\begin{equation}
\label{EQ:INNER}
f[c] = \sum_{m=1}^{M}h_{m,c}^{\rm{T}}x_{m}~ ( c=1,\cdots,C),
\end{equation}
where $\{h_{1,c},h_{2,c},\cdots,h_{M,c}\}$ is the CFB for class $c$; $x_{m}$ is the raw feature vector of the $m$-th face subregion; $C$ is the dimension of the global feature vector (which is equal to the number of face classes in the training set) and $M$ is the number of face subregions in a face image.
% that are represented in terms of space domain since represent the inner products of the local f4eatures and the MS-CFB.
%derived by the addition of the inner products of the local features and the MS-CFB that are represented in terms of space domain.
%The inner product of the raw feature vector and the correlation filter in the MS-CFB (The detailed derivation of CFB is given in Section 2.3) is computed.

\subsection{Classification Rule}
% Why not Maximum Value
After the feature extraction step for both training set and test set, we need to design a classifier for final classification.
 Note that the design process of a CFB is to produce a correlation peak only for the authentic samples for the class of interest, which means that the maximal value criterion,  i.e., the class index of the maximal component in the feature vector, can be used as the classification rule. Thus the label of a test sample can be given according to
\begin{equation}
\label{EQ:12}
Label(y) =   \rm{arg} \max_{\emph{i}=\emph{1},\cdots,\emph{C}}(\emph{y}[\emph{i}]),
\end{equation}
where $\emph{y}=(\emph{y}[1],\emph{y}[2],\cdots,\emph{y}[C])^{\rm{T}}$ is the extracted feature vector corresponding to the test face image.

On the other hand, the cosine similarity measure based nearest neighbor classifier can also be employed for classification. The cosine similarity measure is shown as follows:
\begin{equation}
\label{EQ:11}
Cos(y_{1},y_{2}) =  \frac{y_{1}^{\rm{T}}y_{2}}{||y_{1}||\cdot||y_{2}||},
\end{equation}
where $||\cdot||$ represents the $L_2$ norm. The cosine similarity measure calculates the angle between two vectors and is not affected by their magnitudes.

The cosine similarity measure based nearest neighbor classifier is widely used in face recognition \cite{Klare2013,Nguyen2011}. In \cite{Kumar2006}, it has been shown that the cosine similarity measure performs better than both $L_{1}$ norm and $L_{2}$ norm distance measures in most face recognition experiments. One reason is that \cite{Nguyen2011}, when an unseen sample in the test set is projected onto the feature space, the novel variations in the sample are inclined to evenly affect the projected scale on each component of the features. Thus the variations make more influence on the $L_{1}$ norm and $L_{2}$ norm distance measures (since they are affected by the scale differences \cite{Liu2006}) rather than the angle between two vectors (i.e., the cosine similarity measure). Therefore, the cosine similarity measure, which is invariant to changes in scale, is more effective to perform the nearest neighbor search in the feature space for face recognition.

As a matter of fact, compared with the maximal value criterion, the nearest neighbor classifier based on cosine similarity measure has two main advantages: 1) It explores the information in all components of the feature vectors in both training and test sets, which is beneficial for classification; 2) It can be applied to standard face recognition test protocols (such as FERET \cite{Phillips1998} and CAS-PEAL \cite{Gao2008}).
According to these protocols, the subjects in both gallery and probe sets can be the unseen classes (which do not exist in the training set). In such a case, each component in the extracted feature vectors obtained by MS-CFB characterizes the identity similarity between a training class and the unseen classes. Thus, the maximal value criterion is not valid for classifying the unseen classes, while the nearest neighbor classifier (comparing the feature vectors in the gallery and probe sets) can be used.
%, when. Therefore,.
%Therefore, to overcome the above two issues, we use the cosine similarity based nearest neighbor (NN) classifier to perform classification.See Fig. \ref{FIG:Cosine} for an illustration
 %      \begin{figure*}[tbh!]
%         \centering
%         \scalebox{0.85}{
%            \includegraphics[width=12cm,height=10cm]{cosineSimilarity}
%            }
%         \caption{An illustration of the cosine similarity. Suppose there are six classes, (a) and (c) represent the features G$_{1}$ and G$_{2}$ from Class 2 and Class 5 in the training set, respectively. (b) (or (d)) represent the feature F from Class 2 in the testing set.  By using Eq. (12), (b) (or (d)) is misclassified as Class 5 since the class index of the maximal component is Class 5 (the false peak for the testing subject). However, the cosine similarity between G$_{1}$ and F (0.9060) is higher than that between G$_{2}$ and F (0.8109), which is the desirable result. The cosine similarity uses all the information in the feature for classification.
%                 }
%         \label{FIG:Cosine}%
%        \end{figure*}
%where $y$ is the extracted feature by MS-CFB.
\subsection{The Complete Algorithm}
In the previous subsections, we have developed all ingredients for a robust face recognition algorithm. Now we put them together to yield a complete Multi-Subregion based CFB (MS-CFB) algorithm for face recognition (as detailed in Fig.~\ref{FIG:Algorithm}).
%% New added
\begin{figure}[tbh!]
         \centering
\scalebox{0.8}{
\begin{tabular}
{p{365pt}}
\hline
%\vspace{0.05mm}
%\vspace{2mm}
%\hline
\vspace{0.1mm}
%\vspace{0.1mm}
\textbf{Training Stage:}\\
\textbf{Input:} A training data matrix with $C$ classes; the size of a face subre-\\
 \quad \quad \quad~ gion ($s$); and the regularized parameter ($\alpha$).\\
\textbf{Output:} A feature matrix $\rm{Y}_{train}$ of the training data.\\
\emph{Step 1}: Divide all face images into $M$ blocks of the same size and \\
 \quad \quad \quad~  construct the training data matrix $\rm{X}_{train}$ (see Section 3.1);\\
 %by concatenating the features \\
 %\quad \quad \quad~ of all the face subregions (see Section 3.1);\\
\emph{Step 2}: Do for $c$ = 1,$\cdots$, $C$: \\
 \quad \quad \quad ~2.1 Calculate the covariance matrix $\Sigma_{c}$ via Eq.~(\ref{EQ:RC});\\
 %\quad \quad \quad \quad~ samples of class $c$;\\
 \quad \quad \quad ~2.2 Calculate the mean value $m_{c}$  via Eq.~(\ref{EQ:mean}); \\
 %\quad \quad \quad \quad~ ples of class $c$;\\
 \quad \quad \quad ~2.3 Design the correlation filter bank $g_{c} = (h_{1,c};h_{2,c};\cdots;h_{M,c})$\\
   \quad \quad \quad ~~~via Eq.~(\ref{EQ:9});\\
%\emph{Step 3}: Obtain the correlation filter banks for all classes $G=$\\
 %\quad \quad \quad~ $\{g_{1},g_{2},\cdots,g_{C}\}$;\\
\emph{Step 3}: Compute the feature matrix $\rm{Y}_{train}$ based on the sum of the inner \\
 \quad \quad \quad~  products between $\rm{X}_{train}$ and $\{g_{c}|c=1,2,\cdots,C\}$ via Eq.~(\ref{EQ:INNER}).\\
% \quad \quad \quad~ \\

\vspace{0.05mm}
\textbf{Test Stage:}\\
\textbf{Input:} A test image; and a feature matrix $\rm{Y}_{train}$ of the training data.\\
\textbf{Output:} The class label of the test image.\\
\emph{Step 1}:  Divide the test face image into $M$ blocks of the same size and \\
 \quad \quad \quad~  construct the test data $\rm{x_{test}} $ (see Section 3.1);\\
 %by concatenating the features of \\
 %\quad \quad \quad~  all the face subregions (see Section 3.1);\\
\emph{Step 2}: Compute the feature vector $\rm{y}_{test}$  based on the sum of the inner \\
 \quad \quad \quad~ products between $\rm{x}_{test}$ and $\{g_{c}|c=1,2,\cdots,C\}$  via Eq.~(\ref{EQ:INNER});\\
\emph{Step 3}: Assign the class label to the test image by using the nearest  \\
 \quad \quad \quad~  neighbor classifier with the cosine similarity measure based on \\
  \quad \quad \quad~ $\rm{y}_{test}$ and $\rm{Y}_{train}$.\\
\hline
\end{tabular}
}
         \caption{The complete MS-CFB algorithm for face recognition.
                 }
         \label{FIG:Algorithm}%
        \end{figure}

\subsection{Discussion}
The advantages of the proposed MS-CFB algorithm over the related FE algorithms are summarized as follows.~Firstly, different from traditional global-based and local-based FE algorithms, the proposed algorithm can be viewed as a hybrid algorithm, which uses local facial features to extract a global feature vector. Similar to the human perception system, a hybrid algorithm could combine the advantages of both global-based and local-based FE algorithms, and it is more robust to variations of illumination, facial expression, pose, and so on.~Secondly, compared with the existing local-based FE algorithms, where classifiers are independently trained for each face subregion, a CFB is designed by jointly optimizing multiple correlation filters corresponding to respective face subregions at the overall origin correlation outputs. Therefore, the differences among face subregions are taken into account and the discriminative information in face subregions is more effectively exploited in MS-CFB.~Thirdly,
while the local FE step and the combination of local subregions are considered as two independent processes in traditional local-based FE algorithms, the proposed algorithm attempts to unify these two processes in one framework, where the local FE steps for different face subregions are integrated to produce the optimal outputs. Hence, the effectiveness of local FE is enhanced.
%Fourth, the traditional CFA algorithms \cite{Kumar2006,Yan2008} are based on the signals in terms of Fourier transforms, MS-CFB %directly uses the original raw feature, which reduces the complexity by removing the Fourier transforms.

It is worth mentioning that a CFB becomes an unconstrained correlation filter when a whole face image without division (i.e., $M=1$) is considered. Compared with the constrained correlation filters, such as OTF \cite{Kumar2006,Kumar1999} and OEOTF \cite{Yan2008}, the generalization capability of the unconstrained correlation filter is greatly improved since the hard constraints are removed during the filter design process.
In fact, a CFB with $M=1$ can be viewed as an unconstrained extension of an OEOTF which concentrates on the origin peaks. However, the main differences between a CFB and an OEOTF are: 1) A CFB is designed based on the spatial domain while an OEOTF is represented in the frequency domain. Therefore, traditional Fourier transforms are not required during the design process of a CFB; 2) Compared with an OEOTF that is a single filter, a CFB consists of multiple filters corresponding to different face subregions. A CFB is more robust in dealing with pose variations (by dividing a whole face image into multiple subregions) than an OEOTF.

\section{Experiments}
In this section, we present extensive experimental results on various public face databases to evaluate the effectiveness of the proposed algorithm. In Section 4.1, we introduce the competing algorithms and experimental settings. In Section 4.2, we give the determination of the optimal parameters in MS-CFB.
In Section 4.3, we demonstrate the robustness of the proposed MS-CFB algorithm against illumination variations on the Multi-PIE and FRGC face databases. In Section 4.4, we evaluate the proposed MS-CFB algorithm against pose and facial expression variations on the FERET and LFW face databases. In Section 4.5,  the face recognition performance obtained by the competing algorithms on the databases with a single sample per person is presented. A comprehensive evaluation on the CAS-PEAL R1 face database is shown in Section 4.6. \textcolor{red}{The computational complexity of the proposed algorithm and the performance of the competing algorithms for automatic face recognition are given in Sections 4.7 and 4.8, respectively.} Finally, the discussion is given in Section 4.9.

\subsection{The Competing Algorithms and Experimental Settings}

%\subsubsection{The Competing Algorithms}
To evaluate the performance of the proposed algorithm, we select several popular algorithms for comparisons, including the baseline Eigenface \cite{Turk1991}, Fisherface \cite{Belhumeur1997}, OTF-based \cite{Kumar2006} and OEOTF-based \cite{Yan2008} CFA , Sparse Representation based Classification (SRC) \cite{Wright2009}, and the state-of-the-art local-based FE algorithms including Block-FLD \cite{ Chen2004}, Cascaded LDA (C-LDA) \cite{Kim2005}, Hierarchical Ensemble Classifier (HEC) \cite{Su2009}, Block-based Bag-Of-Words (BBOW) \cite{Li2010}, and Patch-based Collaborative Representation based Classification (PCRC) \cite{Zhu2012}.

%\subsubsection{Experimental Settings}
Each image in the face databases is normalized to extract a facial region that contains only the face. Specifically, the normalization for each image contains the following steps: firstly, the centers of the eyes are manually annotated; secondly, rotation and scaling transformations align the centers of the eyes to predefined locations and fixed interocular distances; finally, a face image is cropped and resized to the size of $80 \times 88$ pixels. Histogram equalization is then applied to all face images for photometric normalization.

The reduced dimension of the PCA subspace in CFA is set to $N-1$, where $N$ is the number of training samples. The value of the parameter $\lambda$ in SRC is set to 0.001 (which is the same as \cite{Zhu2012}). For Block-FLD, we test three different sizes of a face subregion (i.e., $10\times10$, $20\times20$, and $30\times30$) and report the best recognition results obtained with the size of $20\times20$. For C-LDA, the five components encoding scheme is used. For HEC, the size of a candidate face subregion is set to a range from $16\times16$ to $64\times64$. For PCRC, the size of a face subregion is set to $10 \times 10$. For other parameters used in the competing algorithms, we use their default parameter settings.

After feature extraction for both training set and test set, we employ the nearest neighbor classifier for final classification. The cosine similarity measure is used for all compared algorithms. For the proposed MS-CFB algorithm, we respectively evaluate the MS-CFB (max) method (using the maximal value criterion for classification) and the MS-CFB (cos) method (using the cosine similarity measure based nearest neighbor classifier).

%Therefore, for all the evaluation algorithms, the cosine similarity is employed for the NN classifier, expressed as \begin{equation}
%\label{EQ:11}
%Cos(y_{1},y_{2}) =  \frac{y_{1}^{\rm{T}}y_{2}}{||y_{1}||\cdot||y_{2}||},
%\end{equation}
%where $||\cdot||$ represents the $L_2$ norm.
%
%
%
% Note that the design process of CFB is to produce a correlation peak only for the authentic samples for a class of interest, which means that the maximal value criterion,  i.e., the class index of the maximal component in the feature vector, can also be used to perform classification. Thus the label of a test sample can be given according to
%\begin{equation}
%\label{EQ:12}
%Label(y) =   \rm{arg} \max_{\emph{i}=\emph{1},\cdots,\emph{C}}(\emph{y}[\emph{i}]),
%\end{equation}
%where $\emph{y}=(\emph{y}[1],\emph{y}[2],\cdots,\emph{y}[C])^{\rm{T}}$ is the extracted feature vector of a test face image.
%Therefore, for the proposed MS-CFB algorithm, we respectively evaluate the MS-CFB (cos) method (using the cosine similarity measure for the NN classifier) and the MS-CFB (max) method (using the maximal value criterion for classification).

For all databases, a random subset (with $t$ images per subject) is taken from each database to form the training set. The rest of the database is used as the test set.
For each $t$, the experiments with randomly chosen subsets are performed twenty times. We report the average recognition rates as well as the standard
deviations over the randomly chosen test sets as the final results. \textcolor{red}{The training set and test set for all the
competing algorithms are the same for all the experiments.} In addition, the highest recognition rate
for each case is shown in bold font.

In this paper, we focus on the SSS problem, which is one of the most challenging issues in face recognition \cite{Tan2006,Zhu2012}. This problem arises when the number of the samples is smaller than the dimension of the facial feature space. In many real-world applications, the number of training samples for each subject is very limited. Therefore, the discriminability of features under such a case is important to the final performance of a face recognition algorithm.  \textcolor{red}{To evaluate the effectiveness of different feature extraction algorithms to solve the SSS problem, the value of $t$ is set to $2 \sim 5$ for all databases. In Section 4.5, we will discuss the case that the value of $t$ is set to 1 for the SSPP problem in particular.}

% so that the evaluation of the capability of to learn an effective feature extraction is of importance in such scenarios. It is necessary to test such cases.  Therefore,

\subsection{Determining the Optimal Parameters in MS-CFB}
In MS-CFB, two parameters (i.e., the size of a face subregion $s$ and the regularized parameter $\alpha$) have an influence on the recognition accuracy. If the size of a face subregion is too large (e.g., it contains the whole face region), MS-CFB does not take advantage of local-based feature extraction. On the contrary, if the size of a face subregion is too small, MS-CFB becomes sensitive to face alignment. Similarly, the regularized parameter should also be carefully set. The purpose of regularization is to reduce the high variance related to the estimation of the covariance matrix \cite{Lu2005}, which is caused by the SSS problem.

\begin{figure}[tbh!]
         \centering
            \includegraphics[width=12cm,height=3.5cm]{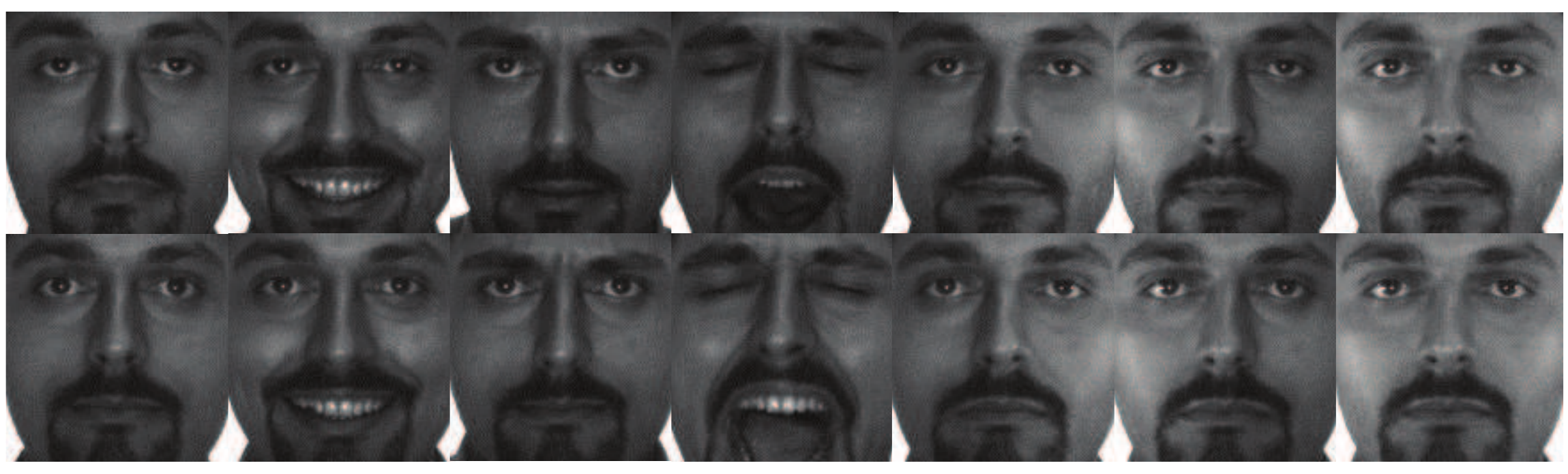}%
         \caption{ The face images of one subject on the AR database.
                 }
         \label{FIG:AR}%
        \end{figure}

 To determine the optimal values of these two parameters (i.e., $s$ and $\alpha$) for MS-CFB, we use the AR database \cite{Martinez1998} for evaluation. The AR database contains over 4,000 face images of 126 subjects (70 men and 56 women). The AR database characterizes the divergence from ideal conditions by incorporating various facial expressions (neutral, smile, and scream), illumination changes (left light on,  right light on, and both sides' light on), and occlusion modes. It has been used as a testbed to evaluate the face recognition algorithms. \textcolor{red}{A subset that contains 120 subjects (each subject has 14 images) with only facial expression and illumination changes is used in our experiments (see Fig.~\ref{FIG:AR} for some examples).}

         \begin{figure}[tbh!]
         \centering
         \subfigure[$t=2$]{
            \includegraphics[width=6.5cm,height=4.5cm]{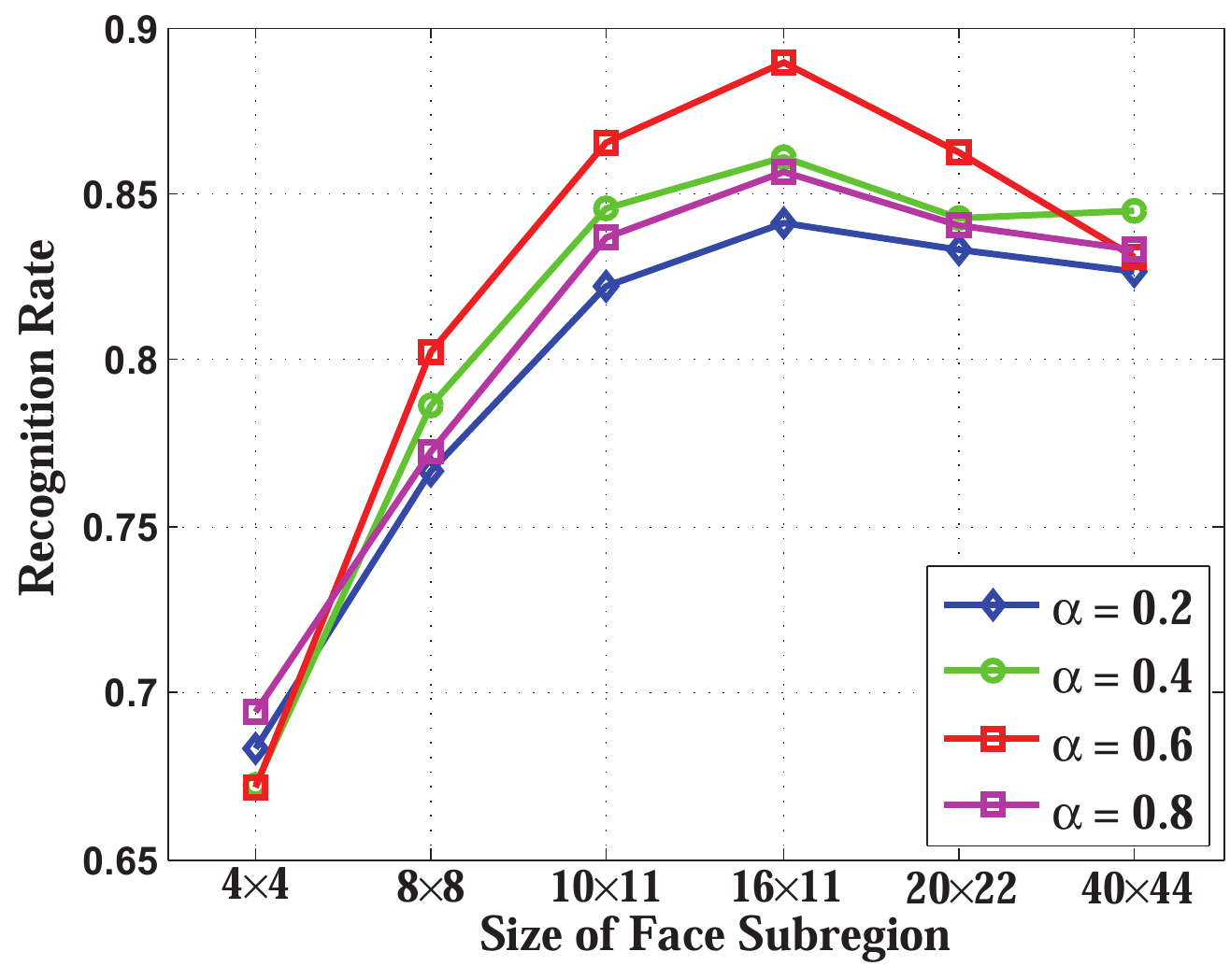}%
            }
          \subfigure[$t=4$]{
            \includegraphics[width=6.5cm,height=4.5cm]{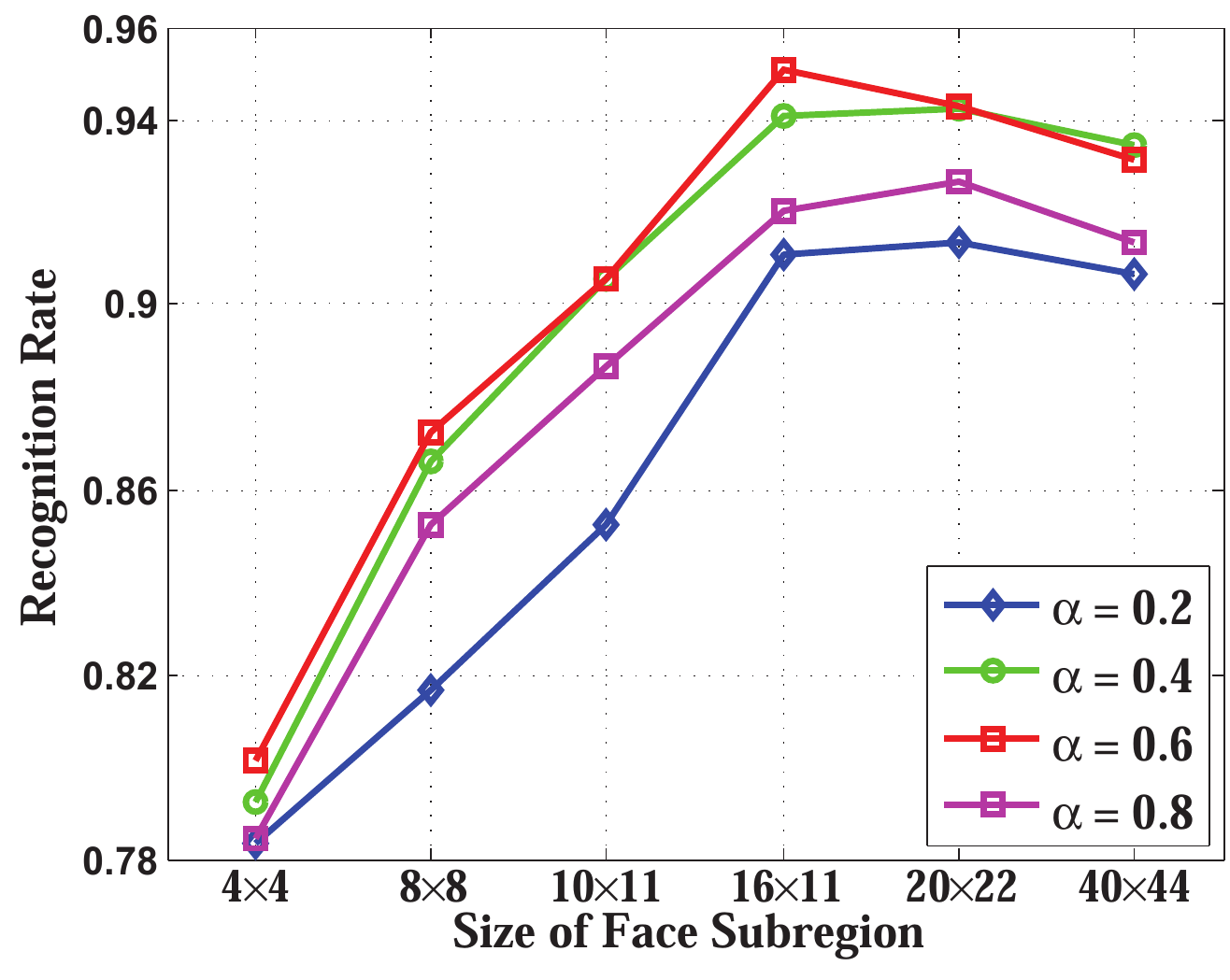}%
            }
         \caption{ The recognition rates obtained by MS-CFB over different sizes of a face subregion and different values of $\alpha$ under $t=2$ and $t=4$ on the AR database.
                 }
         \label{FIG:Tradeoff}%
        \end{figure} %50 male subjects and 70 female subjects

  Fig.~\ref{FIG:Tradeoff} shows the recognition rates obtained by MS-CFB (with the cosine similarity measure) over different sizes of a face subregion (including $4\times4$, $8\times8$, $10\times11$, $16\times11$, $20\times22$, and $40\times44$) and different values of $\alpha$ (including 0.2, 0.4, 0.6, and 0.8) under $t=2$ and $t=4$ on the AR database. We can observe that when the size of a face subregion is very small (e.g., $4\times4$), the recognition rate is low. This is because that a face region is divided into too many subregions, which over-segments meaningful facial features (such as eyes and nose) that are critical for recognition. The recognition rates increase when the size of a face subregion becomes larger. The recognition rate achieves the highest when the size of a face subregion is $16\times11$, while the recognition rate begins to decrease for larger subregion sizes (e.g., $20\times22$ and $40\times44$), which is caused by the sensitivity of large face subregions to variations of facial expression and illumination. The value of the regularized parameter $\alpha$ also affects the recognition accuracy of MS-CFB. When $\alpha = 0.6$, MS-CFB achieves the best results compared with the other values of $\alpha$. Therefore, we choose the size of a face subregion to be $16\times11$ and the value of $\alpha$ to be 0.6 for MS-CFB in all following experiments.

%Table \ref{tab:AR} shows the performance of the competing algorithms and our MS-CFB algorithm under different values of $t$ on the AR database.
%It can be clearly seen that MS-CFB (cos) achieves the highest recognition rates with the training sample size increasing from 2 to 5. In comparison, MS-CFB (max) obtains worse results than MS-CFB (cos). This is because that the maximal value criterion discards the information in other components of the feature vector which is beneficial for classification.
%\textcolor{red}{Moreover, the local-based FE algorithms, such as C-LDA, HEC, BBOW, and PCRC, achieve better results than the global-based FE algorithms (i.e., Eigenface, Fisherface, and SRC), which validates the effectiveness of local facial features for face recognition.}
%We also note that the recognition rates of CFA and SRC \cite{Yan2008,Zhu2012} are low when $t$ is small (e.g., $t=2$). This is due to the fact that more training samples are required to boost the performance of these two algorithms.

\subsection{Robustness to Illumination Variations}

One of the most fundamental challenges in face recognition is significant facial appearance variations due to illumination
changes. In this section, we evaluate the performance of the proposed algorithm against illumination variations on two popular face databases, i.e., the Multi-PIE database \cite{Sim2002} and the FRGC database \cite{Phillips2005}.

\begin{figure*}[tbh!]
         \centering
            \includegraphics[width=14 cm,height=3.5cm]{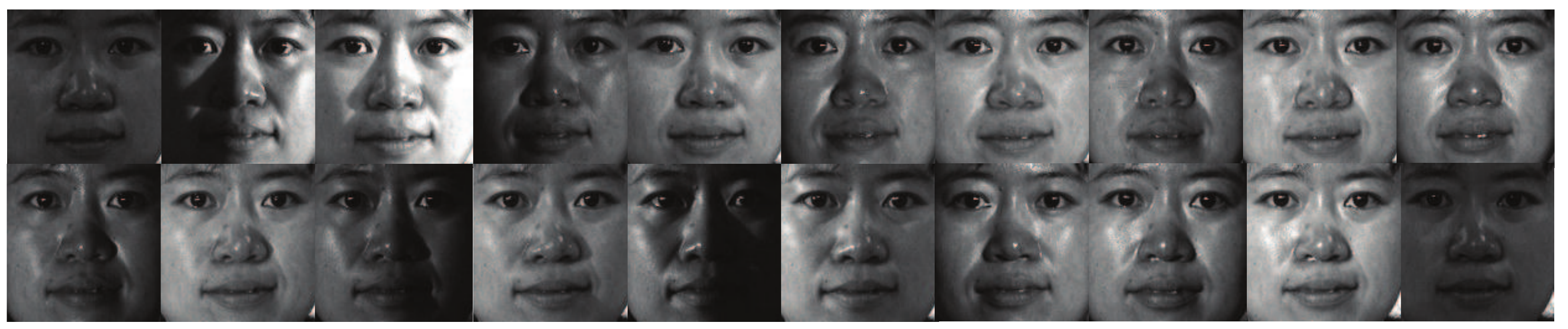}%
         \caption{ The face images of one subject on the Multi-PIE database.
                 }
         \label{FIG:PIE}%
        \end{figure*}

\begin{figure*}[tbh!]
         \centering
            \includegraphics[width=14 cm,height=3.5cm]{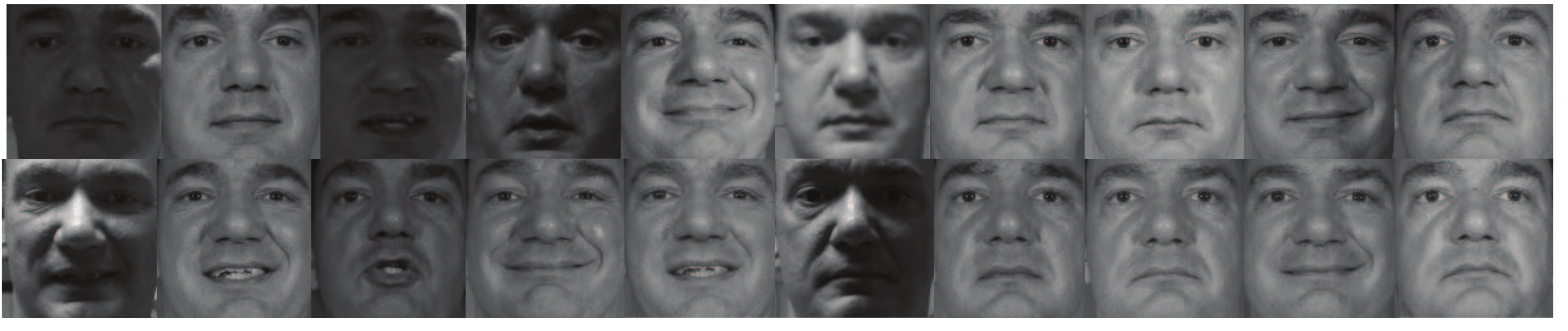}%
         \caption{ The face images of one subject on the FRGC database.
                 }
         \label{FIG:FRGC}%
        \end{figure*}

The Multi-PIE database contains more than 750,000 images of 337 subjects captured in four sessions with variations in pose, facial expression, and illumination. A subset that contains 68 subjects (each subject has 22 images) with various illumination changes is used.  \textcolor{red}{Specifically, we use the frontal pose images (i.e., the c27 subset) under 11 different illumination conditions (i.e., f01, f03, f05, f07, f09, f11, f13, f15, f17, f19, f21) with the ambient lights on/off.
 Fig.~\ref{FIG:PIE} shows the face images of one subject on the Multi-PIE database.}
The FRGC (Face Recognition Grand Challenge) database consists of controlled images, uncontrolled images and three-dimensional images for each subject. We select a subset containing 6,000 images of 300 subjects (20 images for each subject) from the FRGC database. The face images in this subset are captured in both controlled and
uncontrolled conditions with severe illumination variations.  \textcolor{red}{Fig.~\ref{FIG:FRGC} shows the face images of one subject on the FRGC database used in our experiments.}
%        \begin{figure*}[tbh!]
%         \centering
%            \includegraphics[width=7.6cm,height=4cm]{PIE}%
%         \caption{ Face images of the same person on the PIE database.
%                 }
%         \label{FIG:PIE}%
%        \end{figure*}
%
%       \begin{figure*}[tbh!]
%         \centering
%            \includegraphics[width=7.6cm,height=4cm]{FRGC}%
%         \caption{ Face images of the same person on the FRGC database.
%                 }
%         \label{FIG:FRGC}%
%        \end{figure*}
%        \begin{figure*}[tbh!]
%         \centering
%         \subfigure[Multi-PIE]{
%            \includegraphics[width=6cm,height=1.5cm]{PIE}%
%            }
%          \subfigure[FRGC]{
%            \includegraphics[width=6cm,height=1.5cm]{FRGC}%
%            }
%         \caption{The face images of two subjects on the (a) Multi-PIE and (b) FRGC databases.
%                 }
%         \label{FIG:PIEFRGC}%
%        \end{figure*}

%-----------------------------------------------------------
%table:
         \begin{table}
         \centering
         %\finegap
         \caption
         {
          The average recognition accuracies (mean\%$\pm$std.dev.) obtained by the different algorithms on the Multi-PIE database.
         }
         \scalebox{0.8}{
         \begin{tabular}{c|cccc}
         %&\multicolumn{4}{|c|}{Lowest Error Rate}\\%Lowest Error Rate & & & & & & & & & & \\
         \toprule
         Algorithm &$t=2$	&$t=3$ 	&$t=4$	&$t=5$\\
         \midrule
          Eigenface  &72.24$\pm$1.5 &78.54$\pm$1.7 &82.13$\pm$1.8 &85.41$\pm$1.7\\
          \hline
         Fisherface  &76.79$\pm$1.1 &86.63$\pm$1.3 &88.95	$\pm$1.4&92.07$\pm$1.5\\
         \hline
          CFA (OTF)  &83.15$\pm$0.8 &88.05$\pm$0.8 &90.17$\pm$0.6	 &93.10$\pm$0.5 \\
          \hline
          CFA (OEOTF)  &84.00 $\pm$0.6 &88.10$\pm$0.9 &92.32$\pm$0.5	 &93.58$\pm$0.6 \\
          %1D-CFA  &83.65 &92.40 &62.98 &74.95 \\
          %(OEOTF) &(200) &(200) &(300) &(300) \\
         \hline
          SRC  &82.24$\pm$1.2 &86.59$\pm$1.3 &89.98 $\pm$1.2&93.15 $\pm$0.9\\
         \hline
          Block-FLD  &81.17 $\pm$1.0 &82.84$\pm$1.2 &88.77$\pm$1.1 &89.73$\pm$1.0 \\%0.9183	 0.9621
          \hline
          C-LDA   &83.25$\pm$0.9 &85.77$\pm$0.8&89.95 $\pm$0.9&90.07$\pm$0.8\\
          \hline
          HEC &85.56 $\pm$0.8&88.74 $\pm$0.6&91.41 $\pm$0.8&91.11 $\pm$0.6\\
          \hline
          BBOW & 83.58$\pm$0.8  & 87.25$\pm$0.9 &  91.27$\pm$0.9 &  92.66$\pm$0.7 \\
          \hline
          PCRC &86.17 $\pm$0.5&90.15 $\pm$0.7&92.17 $\pm$0.6&93.05 $\pm$0.5\\
          \hline
          MS-CFB (max) &82.51$\pm$1.1  &86.24$\pm$0.9 &90.05  $\pm$0.8 &91.17 $\pm$0.6\\
           \hline
          MS-CFB (cos) &\bf{86.87}$\pm$0.6  &\bf{92.07}$\pm$0.7 &\bf{94.17}  $\pm$0.5 &\bf{96.65} $\pm$0.4\\
         \bottomrule
         \end{tabular}
        }
         \label{tab:PIE}
         \end{table}

 %-----------------------------------------------------------
%table:
         \begin{table}
         \centering
         %\finegap
         \caption
         {
          The average recognition accuracies (mean\%$\pm$std.dev.) obtained by the different algorithms on the FRGC database.
         }
         \scalebox{0.8}{
         \begin{tabular}{c|cccc}
         %&\multicolumn{4}{|c|}{Lowest Error Rate}\\%Lowest Error Rate & & & & & & & & & & \\
         \toprule
         Algorithm &$t=2$	&$t=3$ 	&$t=4$	&$t=5$\\
         \midrule
          Eigenface  &45.38$\pm$1.3 &53.10$\pm$1.2 &64.35$\pm$1.1 &70.26$\pm$1.5\\
          \hline
         Fisherface  &48.17$\pm$1.1 &55.42$\pm$1.3 &66.78	$\pm$1.5&69.06$\pm$1.7\\
         \hline
          CFA (OTF)  &54.35$\pm$0.8 &62.17$\pm$0.8 &65.99$\pm$0.9	 &73.81$\pm$1.0 \\
         \hline
          CFA (OEOTF)  &59.80 $\pm$0.7 &70.05 $\pm$0.9 &78.31 $\pm$0.7	 &85.04$\pm$0.6 \\
          %1D-CFA  &83.65 &92.40 &62.98 &74.95 \\
          %(OEOTF) &(200) &(200) &(300) &(300) \\
         \hline
          SRC  &57.72$\pm$1.1 &65.14$\pm$1.2 &72.28 $\pm$0.9&81.18 $\pm$0.9\\
         \hline
          Block-FLD  &53.14 $\pm$0.8 &62.28$\pm$1.3 &66.77$\pm$0.9 &70.20$\pm$1.0 \\%0.9183	 0.9621
          \hline
          C-LDA   &55.72$\pm$1.1 &66.11 $\pm$0.8&72.24 $\pm$1.1&76.89$\pm$1.2\\
          \hline
          HEC &57.28 $\pm$1.3&66.24 $\pm$1.2&71.17 $\pm$1.3&75.25 $\pm$1.5\\
          \hline
          BBOW & 58.57$\pm$1.4  & 71.90$\pm$1.2 &  73.10$\pm$0.7 & 78.43 $\pm$0.9 \\
          \hline
          PCRC &59.02 $\pm$1.0&70.02 $\pm$1.0&75.65 $\pm$0.6&80.11 $\pm$0.5\\
          \hline
          MS-CFB (max) &59.86$\pm$1.2  &70.66$\pm$1.3 &78.31  $\pm$1.2 &85.53 $\pm$1.2\\
          \hline
          MS-CFB (cos) &\bf{63.99}$\pm$0.8  &\bf{75.24}$\pm$0.9 &\bf{82.21}  $\pm$0.5 &\bf{88.58} $\pm$0.6\\
         \bottomrule
         \end{tabular}
        }
         \label{tab:FRGC}
         \end{table}

         Tables \ref{tab:PIE} and \ref{tab:FRGC} show the average recognition accuracies obtained by the different algorithms on the Multi-PIE and FRGC databases, respectively. From these tables, we can see that the proposed MS-CFB (cos) algorithm consistently achieves better recognition accuracies than the other competing algorithms. Compared with MS-CFB (max), MS-CFB (cos) improves the recognition rates by about $4\%\sim5\%$, which demonstrates the advantages of using the cosine similarity measure as a metric. SRC obtains better results than Block-FLD in Multi-PIE and FRGC, which shows that SRC is more robust in dealing with illumination variations. Block-FLD constructs multiple training patterns from a single image, but it does not consider the relationships among different face subregions. PCRC, HEC, and BBOW achieve worse performance than MS-CFB (cos). The reason is that MS-CFB considers the local FE step and the combination of different face subregions as a whole, which effectively overcomes the disadvantages of the conventional fusion strategies (e.g., the majority voting used in PCRC,  the weighted sum of local facial features used in HEC, and the concatenation of local features used in BBOW) employed in local-based FE algorithms.
%\vspace{-0.2cm}
\subsection{Robustness to Pose and Facial Expression Variations}
%\vspace{-0.2cm}
In this section, we evaluate the influence of pose and expression variations on the performance of the proposed algorithm by using two representative face databases, i.e., the FERET database \cite{Phillips1998} and the LFW database \cite{Huang2007}.

\begin{figure*}[tbh!]
         \centering
            \includegraphics[width=10 cm,height=2.2cm]{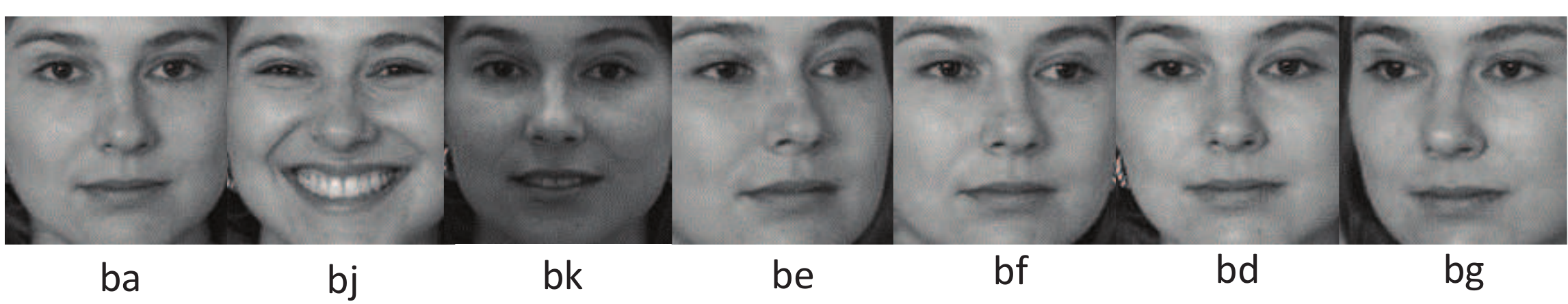}%
         \caption{ The face images of one subject on the FERET database.
                 }
         \label{FIG:FERET}%
        \end{figure*}

\begin{figure*}[tbh!]
         \centering
            \includegraphics[width=14 cm,height=1.8cm]{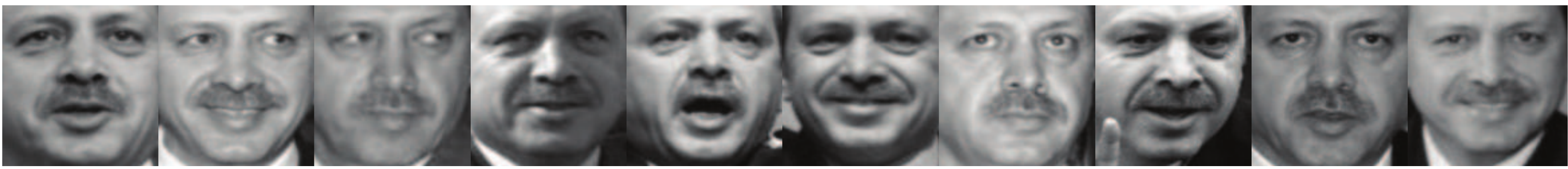}%
         \caption{ The face images of one subject on the LFW database.
                 }
         \label{FIG:LFW}%
        \end{figure*}

The FERET database is a standard face database for evaluating the performance of face recognition algorithms. \textcolor{red}{A subset of the FERET database, which includes 1,400 images of 200 subjects (each subject has 7 images), is used.
%This subset involves challenges, such as variations in facial expression and pose. Fig.~\ref{FIG:FERET} shows the sample images of two subjects on the FERET and %LFW databases used in our experiments.
It is composed of the images whose names are marked with two-character strings: ``ba", ``bj", ``bk", ``be", ``bf", ``bd", and ``bg" (see \cite{Phillips1998} for more details), as shown in Fig.~\ref{FIG:FERET}. This subset involves challenges, such as variations in facial expression and pose.}
Besides, we also perform
an experiment on a more realistic face database captured in unconstrained environments (i.e., the Labeled Faces in the Wild (LFW) database). The LFW database is usually used to evaluate face recognition algorithms in real scenarios. It contains the images of 5,749 different individuals collected from the web. LFW-a \cite{Wolf2011} is a version of LFW after face alignment. A subset with 150 subjects (10
images for each subject) is chosen from LFW-a. This subset
involves severe variations in pose, facial expression, etc.  \textcolor{red}{Fig.~\ref{FIG:LFW} shows the sample images of one subject on the LFW database used in our experiments.}

%\begin{figure*}[tbh!]
%         \centering
%         \subfigure[FERET]{
%            \includegraphics[width=6cm,height=1.5cm]{FERET}%
%            }
%          \subfigure[LFW]{
%            \includegraphics[width=6cm,height=1.5cm]{LFW}%
%            }
%         \caption{The face images of two subjects on the (a) FERET and (b) LFW databases.
%                 }
%         \label{FIG:FERETLFW}%
%        \end{figure*}
%        \begin{figure*}[tbh!]
%         \centering
%            \includegraphics[width=7.6cm,height=4cm]{LFW}%
%         \caption{ Face images of the same person on the LFW database.
%                 }
%         \label{FIG:LFW}%
%        \end{figure*}
%
%       \begin{figure*}[tbh!]
%         \centering
%            \includegraphics[width=7.6cm,height=4cm]{FERET}%
%         \caption{ Face images of the same person on the FERET database.
%                 }
%         \label{FIG:FERET}%
%        \end{figure*}
   %table:
         \begin{table}
         \centering
         %\finegap
         \caption
         {
          The average recognition accuracies (mean\%$\pm$std.dev.) obtained by the different algorithms on the FERET database.
         }
         \scalebox{0.8}{
         \begin{tabular}{c|cccc}
         %&\multicolumn{4}{|c|}{Lowest Error Rate}\\%Lowest Error Rate & & & & & & & & & & \\
         \toprule
         Algorithm &$t=2$	&$t=3$ 	&$t=4$	&$t=5$\\
         \midrule
          Eigenface  &53.27$\pm$3.0 &60.12$\pm$2.9 &65.50$\pm$2.7 &70.22$\pm$2.1\\
          \hline
         Fisherface  &66.63$\pm$1.8 &67.79$\pm$1.7 &76.23	$\pm$1.6&77.54$\pm$1.3\\
         \hline
          CFA (OTF)  &58.96$\pm$1.7 &65.53$\pm$1.5 &74.18$\pm$1.1	 &78.97$\pm$1.4 \\
             \hline
          CFA (OEOTF)  &75.27$\pm$1.5 &79.92$\pm$1.6 &90.02$\pm$1.3	 &91.50$\pm$1.3 \\
          %1D-CFA  &83.65 &92.40 &62.98 &74.95 \\
          %(OEOTF) &(200) &(200) &(300) &(300) \\
         \hline
          SRC  &66.21$\pm$2.1 &67.14$\pm$2.2 &71.16 $\pm$2.5&75.36 $\pm$2.1\\
         \hline
          Block-FLD  &67.57 $\pm$1.8 &69.95$\pm$1.7 &73.28$\pm$1.7 &80.95$\pm$1.6 \\%0.9183	 0.9621
          \hline
          C-LDA   &68.83$\pm$2.1 &70.17 $\pm$2.3 &75.36 $\pm$2.4&83.27$\pm$2.3\\
          \hline
          HEC &71.72 $\pm$1.8&74.92 $\pm$1.7&80.38 $\pm$1.8&85.50 $\pm$1.9\\
          \hline
          BBOW & 74.15$\pm$1.6  &77.42$\pm$1.2 &  86.00$\pm$1.5 & 92.34$\pm$1.5 \\
          \hline
          PCRC &75.24 $\pm$1.5&79.17 $\pm$1.2&87.93 $\pm$1.4&\bf{95.85} $\pm$1.3\\
          \hline
          MS-CFB (max) &75.10$\pm$1.9  &81.14$\pm$1.8 &90.15  $\pm$1.1 &92.11 $\pm$1.4\\
          \hline
          MS-CFB (cos)&\bf{80.60}$\pm$1.4  &\bf{84.72}$\pm$1.3 &\bf{94.26}  $\pm$1.2 &94.93 $\pm$1.1\\
         \bottomrule
         \end{tabular}
        }
         \label{tab:FERET}
         \end{table}

         \begin{table}
         \centering
         %\finegap
         \caption
         {
          The average recognition accuracies (mean\%$\pm$std.dev.) obtained by the different algorithms on the LFW database.
         }
         \scalebox{0.8}{
         \begin{tabular}{c|cccc}
         %&\multicolumn{4}{|c|}{Lowest Error Rate}\\%Lowest Error Rate & & & & & & & & & & \\
         \toprule
         Algorithm &$t=2$	&$t=3$ 	&$t=4$	&$t=5$\\
         \midrule
          Eigenface  &24.15$\pm$3.2 &28.10$\pm$3.8 &32.23$\pm$3.5 &37.00$\pm$3.7\\
          \hline
         Fisherface  &27.89$\pm$2.8 &33.42$\pm$2.7 &38.42	$\pm$2.4&44.25$\pm$2.3\\
         \hline
          CFA (OTF)  &25.27$\pm$3.5 &30.17$\pm$3.9 &32.17$\pm$4.0	 &35.24$\pm$3.5 \\
          %1D-CFA  &83.65 &92.40 &62.98 &74.95 \\
          %(OEOTF) &(200) &(200) &(300) &(300) \\
         \hline
          CFA (OEOTF)  &30.11$\pm$2.1 &35.39$\pm$1.8 &39.95$\pm$1.6	 &42.13$\pm$1.5 \\
         \hline
          SRC  &30.25$\pm$2.5 &35.24$\pm$2.3 &39.97 $\pm$2.8&45.13 $\pm$2.0\\
         \hline
          Block-FLD  &32.53 $\pm$2.3 &36.78$\pm$2.4 &40.12$\pm$1.9 &45.24$\pm$1.5 \\%0.9183	 0.9621
          \hline
          C-LDA   &31.10$\pm$2.2 &35.41 $\pm$2.1&38.82 $\pm$1.5&44.99$\pm$1.3\\
          \hline
          HEC &33.24 $\pm$2.3&41.78 $\pm$2.2&45.80 $\pm$1.5&49.72 $\pm$1.9\\
          \hline
          BBOW & 31.27$\pm$2.2  &  33.41$\pm$1.9 & 41.17$\pm$1.5 & 48.21$\pm$1.5 \\
          \hline
          PCRC &\bf{38.20} $\pm$2.0 &42.17 $\pm$1.4&\bf{48.58}$\pm$1.3&50.72 $\pm$1.3\\
          \hline
          MS-CFB (max) &31.10$\pm$2.4  &35.22$\pm$2.1 &42.32  $\pm$2.0 &46.00 $\pm$1.8\\
          \hline
          MS-CFB (cos) &37.17$\pm$1.8  &\bf{43.10}$\pm$1.5 &47.15  $\pm$1.4 &\bf{52.20} $\pm$1.2\\
          \bottomrule
         \end{tabular}
        }
         \label{tab:LFW}
         \end{table}
%AR

Tables \ref{tab:FERET} and \ref{tab:LFW} show the experimental results on the FERET and LFW databases, respectively. MS-CFB (cos) obtains comparable or better recognition rates than the other algorithms. Particularly, the performance of MS-CFB (cos) increases significantly when more training sample are used. MS-CFB (cos) improves the discriminability of features by adopting the unconstrained form (which is beneficial for learning the underlying classification boundary) during the design process of a CFB.
The recognition accuracies obtained by CFA (OTF) and CFA (OEOTF) are lower than those obtained by MS-CFB (cos).
        \begin{figure*}[tbh!]
         \centering
         \subfigure[]{
            \includegraphics[width=2.5cm,height=2.5cm]{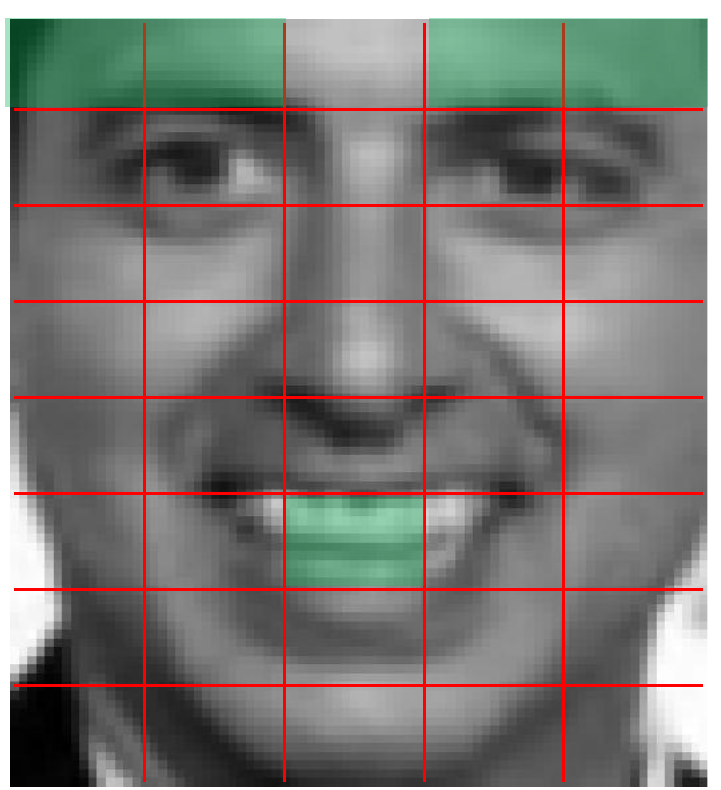}%
            }
          \subfigure[]{
            \includegraphics[width=2.5cm,height=2.5cm]{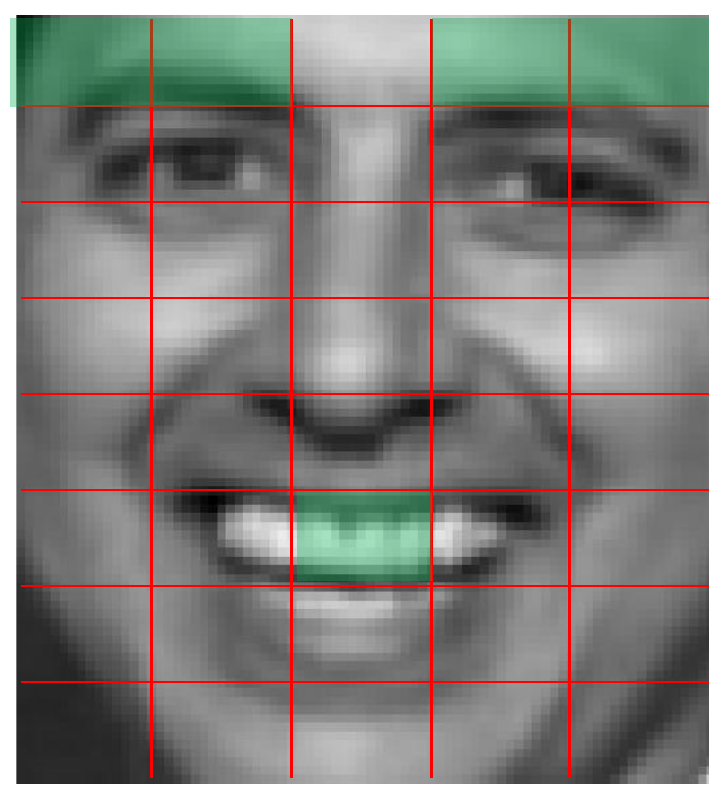}%
            }
          \subfigure[]{
            \includegraphics[width=2.5cm,height=2.5cm]{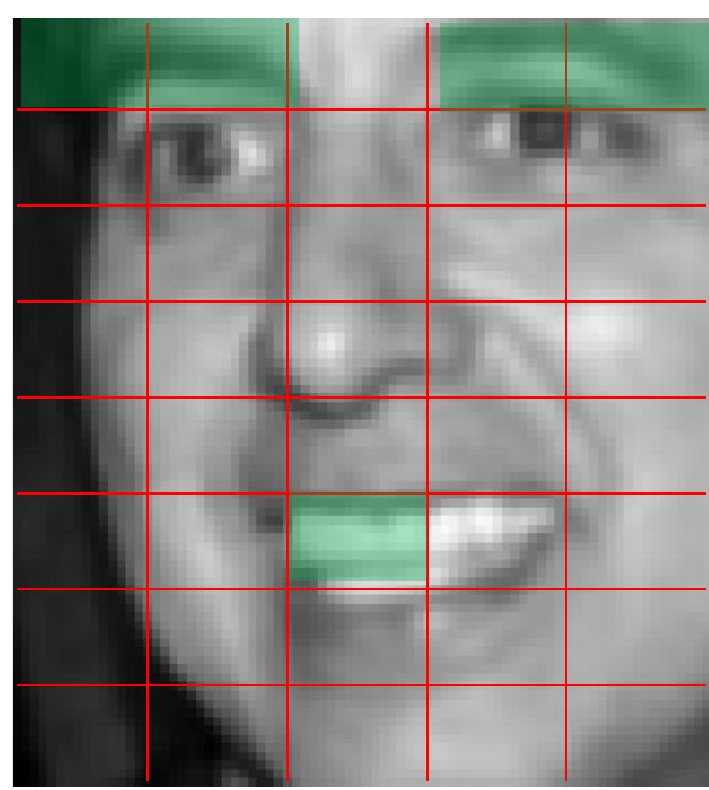}%
            }
         \caption{ Multi-block division of the same subject with different poses based on our face alignment (an image is divided into $5\times 8$ blocks and the size of each block is $16\times11$ pixels). (a) and (b) are both frontal face images while (c) is a face image with a large pose variation. Note that the blocks marked with green in (a) and (b) are aligned, while the marked blocks in (c) are not well-aligned with the ones in (a) and (b).}
         \label{FIG:POSE}%
         \end{figure*}
This is due to the fact that the usage of the whole face region makes CFA sensitive to pose variations. In contrast, MS-CFB (cos) alleviates this problem by using multiple face subregions. Furthermore, BBOW obtains lower recognition rates than HEC and PCRC on the LFW database, which indicates that BBOW cannot effectively capture the intrinsic discriminative information when the training set contains variations in pose and facial expression.

Compared with the recognition results on other databases, MS-CFB (cos) obtains lower accuracies on the LFW database. There are two main reasons: 1) After the multi-subregion division procedure, some face subregions contain the surrounding background (mainly caused by pose changes), which decreases the discriminability of features extracted by our algorithm (note that MS-CFB is based on the sum of the correlation outputs from all face subregions); 2) The mismatching of face subregions between training samples and test samples can occur when dealing with large pose variations. See Fig.~\ref{FIG:POSE} for an example. In our experiments all face images are aligned only according to the manually annotated eye positions, as in \cite{Yan2008,Liu2006}. When handling the frontal face images, most face subregions between training samples and test samples, corresponding to specific facial structures (such as eyes, mouth), can be aligned, which makes our algorithm work well. However, when matching face images with large pose variations, the performance of our algorithm drops. This is because the face alignment method employed in our work is not effective enough so that the blocks with the same spatial layout are not well-aligned in this case, which leads to low correlation values between face subregions and the corresponding correlation filters. Therefore, a more effective face alignment technique can improve the performance of our algorithm, especially for handling images with large pose variations.

%Figure shows an example of multi-block division of a same subject with different poses. We can see that a block corresponds to the same face subregions for frontal facial images, while the blocks might be mismatching for a facial image with a large pose variation.
%In other words, when dealing with frontal face recognition, most blocks fit to same and meaningful facial structures (such as eyebrow, eyes, nose and mouth). The facial structure is helpful to improve the final recognition performance. However, when handling the face with large pose variations, mismatching of face subregions will decrease the recognition performance since blocks in two images with the same position might correspond to different facial structures. Therefore, %However, this issue is beyond the scope of this paper.
%Most of the local-based FE algorithms are more robust than the global-based ones in dealing with pose and facial expression variations.

 %since it uses an effective patch based collaborative representation with regularized margin distribution optimization.
%\subsection{Robustness to Occlusion}
%\vspace{-0.2cm}
\subsection{Face Recognition on Databases with a Single Sample Per Person}
%\vspace{-0.2cm}
In this section, we test the performance of the competing algorithms on all above-mentioned databases with a Single training Sample Per Person (SSPP) \cite{Tan2006,Martinez2002} (which is an extreme case of the SSS problem that severely challenges conventional face recognition algorithms). In such a case, supervised learning techniques, such as LDA \cite{Belhumeur1997}, may not be applicable since the intra-subject information cannot be obtained from one training sample. One possible solution is to use a generic training set. For instance, Su et al.~\cite{Su2010} proposed an Adaptive Generic Learning (AGL) algorithm, which is specially designed for solving the SSPP problem by using a generic training set. Kan et al.~\cite{Kan2013} developed an Adaptive Discriminant Analysis (ADA) algorithm, where the within-class scatter matrix of each single sample is inferred by using only a limited number of the nearest neighbors in the generic training set.
Recently, the image partitioning based algorithms become popular for solving the SSPP problem. Lu et al.~\cite{Lu2013} proposed a novel Discriminative Multi-Manifold Analysis (DMMA) algorithm by learning discriminative features from image patches.
Therefore, AGL, ADA and DMMA are employed as the competing algorithms in our experiments. When we evaluate the performance of AGL (or ADA) on one database, all the other databases are used to constitute the generic training set in AGL (or ADA). For the other algorithms, we only use a single sample per person for training. Note that since Fisherface \cite{Belhumeur1997} (based on LDA) cannot deal with the SSPP problem, its performance is not reported in this section.
         %-----------------------------------------------------------
%table:
         \begin{table}
         \centering
         %\finegap
         \caption
         {
           The average recognition accuracies (mean\%$\pm$std.dev.) obtained by the different algorithms for the SSPP problem.
         }
        \scalebox{0.75}{
         \begin{tabular}{c|ccccc}
         %&\multicolumn{4}{|c|}{Lowest Error Rate}\\%Lowest Error Rate & & & & & & & & & & \\
         \toprule
         Algorithm &AR	&Multi-PIE 	&FRGC	&FERET &LFW\\
         \midrule
          Eigenface  &35.77$\pm$3.5 &50.15$\pm$3.5 &22.42$\pm$4.1 &33.70$\pm$3.8 &11.13$\pm$3.8\\
         \hline
          CFA (OTF)  &38.54$\pm$3.4 &55.54$\pm$2.5 &40.17$\pm$3.8	 &31.00$\pm$3.5  &13.21$\pm$2.8\\
          \hline
          CFA (OEOTF)  &53.27$\pm$2.9 &58.10$\pm$2.1 &43.50$\pm$3.1	 &55.27$\pm$3.0  &16.17$\pm$2.6\\
          %1D-CFA  &83.65 &92.40 &62.98 &74.95 \\
          %(OEOTF) &(200) &(200) &(300) &(300) \\
         \hline
          SRC  &45.27$ \pm$3.2&57.89$\pm$1.8 &38.28$\pm$3.3 &43.82$\pm$3.3  &15.26$\pm$2.7\\
         \hline
         Block-FLD  &48.81$ \pm$2.4 &56.17$\pm$1.4 &45.17$\pm$3.0 &50.47$\pm$2.8  &18.78$\pm$2.5\\%0.9183	 0.9621
         \hline
         AGL  &55.41$ \pm$3.4 &60.95$\pm$1.7 &50.20$\pm$2.8 &55.14$\pm$1.3  &15.11$\pm$3.2\\%0.9183	 0.9621
         \hline
         ADA  &60.18$\pm$3.0 &60.16$\pm$1.9 &51.76$\pm$2.7 & 60.11$\pm$62  &19.32$\pm$3.0\\%0.9183	 0.9621
         \hline
         DMMA  & \bf{67.24$\pm$2.0}   &62.55$\pm1.6$ &\bf{53.15$\pm$2.7} &65.24$\pm2.5$  &\bf{22.17$\pm$2.8}\\%0.9183	 0.9621
         \hline
         BBOW & 64.21$\pm$2.5  & 55.98$\pm$ 1.8 &  46.31$\pm$2.7 &  60.52$\pm$ 3.0 &17.37$\pm$3.2\\
         \hline
          PCRC &65.40$\pm$2.3 &61.11$\pm$1.6 &48.94$\pm$3.2 &64.25$\pm$2.2  &22.14$\pm$2.8\\
         \hline
         MS-CFB (max) &61.21$ \pm$2.9 &57.72$\pm$2.5 &45.30$\pm$2.4 &61.78$\pm$2.1 &16.66$\pm$2.2 \\
          \hline
         MS-CFB (cos) &66.13$ \pm$2.2 &\bf{62.81$\pm$1.5} &52.74 $\pm$2.8 &\bf{66.60}$\pm$2.1 &21.15$\pm$2.9 \\
         \bottomrule
         \end{tabular}
         }
         \label{tab:SSPP}
         \end{table}

         Table \ref{tab:SSPP} shows the average recognition accuracies obtained by the competing algorithms in dealing with the SSPP problem. Among the competing algorithms, MS-CFB (cos) obtains comparable results on most databases. Specifically, MS-CFB (cos) outperforms most of the compared local-based algorithms, such as Block-FLD, BBOW, and PCRC.  Furthermore, it obtains comparable performance with the recently proposed DMMA algorithm which considers the local face subregions of each subject as a manifold. The reason why our algorithm is comparable to these state-of-the-art algorithms even if only raw data of local face subregions is used is that our algorithm extracts global features by effectively combining local features in an integrated framework, while others extract local features independently.
         Furthermore, compared with the AGL and ADA algorithms, which additionally use a generic training set, MS-CFB (cos) still achieves better performance, which clearly demonstrates the desirable classification ability of the proposed algorithm.
         It is also interesting to observe that MS-CFB (cos), DMMA, and PCRC obtain better recognition results than AGL and ADA in most databases.

         \textcolor{red}{Note that the results obtained by some competing algorithms (such as DMMA \cite{Lu2013}, PCRC \cite{Zhu2012}, AGL \cite{Su2010},  and ADA \cite{Kan2013}) in our experiments are different from the reported results. This is because that the experimental settings in our paper and the original papers are different. For instance, in the original papers \cite{Lu2013,Zhu2012}, DMMA used the standard FERET evaluation protocol, while PCRC used more than 2 images per person for training. In contrast, for DMMA and PCRC, we only use a single sample per person for training in our paper.
        In addition, in the original papers \cite{Su2010,Kan2013}, AGL (or ADA) uses a generic training set that is similar to the test set. However, when we evaluate the performance of AGL (or ADA) on one database in this paper, all the other databases are used to constitute the generic training set (which is significantly different from the test set) for AGL (or ADA). Hence, the accuracies of AGL and ADA are lower than those reported in the original papers. How to choose a proper and representative generic training set still needs further investigation for AGL and ADA.}
        %  is because the generic training set is different. In the original papers, AGL and ADA use a generic database similar to the test database. However, when we evaluate the performance of AGL (or ADA) on one database, all the other databases are used to constitute the generic training set in AGL (or ADA). Hence, the accuracy of AGL and ADA are much lower than the original papers.}
%
%         %There are two reasons to explain why our method is comparable to these methods even an additional training set is used: 1) MS-CFB extracts features in a class-specific manner while others extract features in a generic way. 2
%         Therefore, how to choose a proper and representative generic training set still needs further investigation for AGL and ADA.

\subsection{Face Recognition on CAS-PEAL R1 with Unseen Subjects}

\begin{figure*}[tbh!]
         \centering
            \includegraphics[width=6 cm,height=3cm]{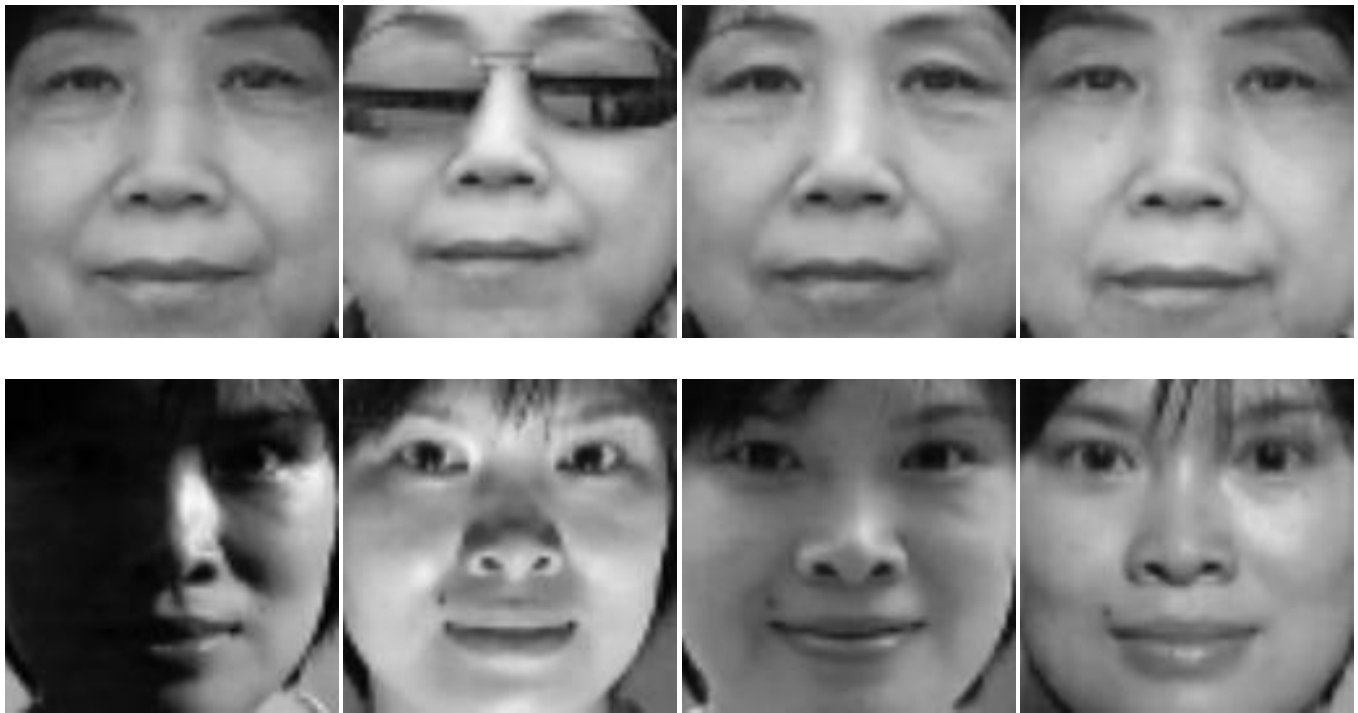}%
         \caption{ The face images of two subjects on the CAS-PEAL R1 training set.
                 }
         \label{FIG:CAS-PEAL}%
        \end{figure*}

To evaluate the generalization capability of the proposed algorithm, we use
the CAS-PEAL R1 face database for evaluation. The CAS-PEAL R1 database contains three types of datasets,
i.e., the training set, gallery set and probe set. The training set contains 300 subjects and each subject has 4 images. The gallery set includes 1,040 images of 1,040 subjects (each subject has one image captured under a normal condition). The CAS-PEAL R1 database contains six probe sets under six different conditions: accessory, age,
background, distance, expression, and lighting. All images that appear
in the training set are excluded from the probe sets and the probe subjects
 may not exist in the training set. We employ the evaluation protocol introduced in \cite{Gao2008}.
 Here only the training set is used to train all of the algorithms.
 The details of the CAS-PEAL R1 database are described in Table \ref{tab:PEALSET}.
 Fig.~\ref{FIG:CAS-PEAL} shows the face images of two subjects on the CAS-PEAL R1 training set.
 Among the competing algorithms, SRC and PCRC are infeasible to deal with the case that the probe subjects are the unseen subjects in the probe sets, because a test image is represented as a linear combination of the training samples for these two algorithms. In addition, MS-CFB (max) is not evaluated, since it is not valid for classifying unseen subjects.
       \begin{table*}[tbh!]
         \centering
         %\finegap
         \caption
         {
           The datasets used in the CAS-PEAL R1 evaluation protocol.
         }
          \scalebox{0.74}{
         \begin{tabular}{c|c|c|cccccc}
         %&\multicolumn{4}{|c|}{Lowest Error Rate}\\%Lowest Error Rate & & & & & & & & & & \\
         \toprule
         \multirow{2}{*}{Datasets} &Training &Gallery &\multicolumn{6}{c}{Probe set (frontal)}\\
         \cline{4-9}
           &set &set &Accessory	&Age 	&Background & Distance & Expression &Lighting \\
         \midrule
         No. of Images &1,200	&1,040	&2,285	&66	&553 &275 &1,570 &2,243\\
         \bottomrule
         \end{tabular}
         }
         \label{tab:PEALSET}
         \end{table*}
        \begin{table*}[tbh!]
         \centering
         %\finegap
         \caption
         {
         The recognition rates (\%) obtained by the different algorithms on the CAS-PEAL R1 database .
         }
         \scalebox{0.76}{
         \begin{tabular}{c|ccccccc}
         %&\multicolumn{4}{|c|}{Lowest Error Rate}\\%Lowest Error Rate & & & & & & & & & & \\
         \toprule
         Algorithm &Accessory	&Age 	&Background & Distance & Expression &Lighting & Average\\
         \midrule
         Eigenface  &59.39 &57.58 &95.84 &93.09 &73.69 &10.16 & 51.00 \\
         \hline
         Fisherface &45.95 &33.33 &87.70 &77.45 &61.34 &4.95 & 40.67 \\
         \hline
         CFA (OTF) &53.52 &56.06 &94.58 &92.00 &67.83 &15.78 & 49.41 \\
         \hline
         CFA (OEOTF) &73.39 &66.67 &\bf{98.19} &\bf{98.18} &83.31 &30.14 & 64.62 \\
          %1D-CFA  &73.39 &66.67 &\bf{98.19} &98.18 &\bf{83.31} &30.14 & \multirow{2}{*}{64.62} \\
          %(OEOTF) &(300) &(300) &\bf{(300)} &(300)  &\bf{(300)} &(300)\\
         %\hline
         %SRC  &60.18 &57.58 &92.59 &94.55 &71.21 &20.33 & 53.76 \\
         \hline
          B-FLD  &65.43 &63.64 &90.60 &93.82 &75.92 &23.09 & 57.29 \\
         \hline
         C-LDA &69.80 &69.70 &94.03 &94.91 &76.05 &24.92& 59.71 \\
         \hline
          HEC &70.68 &71.21 &94.21 &96.36 &82.04 &35.62 &  64.86 \\
         \hline
         BBOW &60.18 &57.58 &92.59 &94.55 &71.21 &20.33 & 53.76\\
         %\hline
         % PCRC &72.43 &71.21 &96.38 &\bf{98.18} &84.71 &38.16 &  67.09\\
         \hline
          MS-CFB (cos) &\bf{75.49} &\bf{75.76} &97.29 &97.82 &\bf{88.28} &\bf{42.71} & \bf{70.45} \\
         \bottomrule
         \end{tabular}
         }
         \label{tab:PEAL}
         \end{table*}

The recognition rates obtained by the different algorithms on the CAS-PEAL R1 database are given in Table \ref{tab:PEAL}.~It can be seen that MS-CFB (cos) achieves the recognition rates with at least 6\% higher (on an average) than the other competing algorithms. Fisherface obtains the worst recognition rates (which are much lower than the recognition rates obtained by Eigenface). The generalization capability of Fisherface is poor because the number of training samples for each class is small. BBOW obtains much worse performance than HEC and C-LDA. The reason is that the codewords learned in the training set are not representative (note that some subjects in the probe sets are different from those in the training set).
MS-CFB (cos) achieves the highest recognition rates on the `\emph{Accessory}', `\emph{Age}', and `\emph{Expression}'  probe sets. In particular, for the most difficult `\emph{Lighting}' probe set, MS-CFB (cos) significantly improves the recognition accuracy (it achieves the recognition rate of 70.45\%), while Fisherface only obtains the recognition rate of 4.95\%. In short, these experimental results on the CAS-PEAL R1 database show that the CFBs learned on the training set can classify unseen subjects well in the proposed MS-CFB.
\subsection{Computational Complexity of the Proposed Algorithm}
We compare the computational time of the proposed MS-CFB algorithm with that of some representative feature extraction algorithms, including Eigenface, Fisherface, CFA (with OTF and OEOTF), and PCRC. All the computational time is reported on a workstation with 2 Intel Xeon E5620 (2.40GHz) CPUs (only one core is used) on the MATLAB platform. Table \ref{tab:TIME} shows the computational time spent on the training and test (recognition) stages by these algorithms on the CAS-PEAL R1 database.
       \begin{table*}[tbh!]
         \centering
         %\finegap
         \caption
         {
           Comparisons of the computational time (in seconds) used by the competing algorithms on the CAS-PEAL R1 database.
         }
         \scalebox{0.87}{
         \begin{tabular}{c|c|c}
         %&\multicolumn{4}{|c|}{Lowest Error Rate}\\%Lowest Error Rate & & & & & & & & & & \\

         \toprule
           Algorithm &Training time	 &Recognition time \\
         \midrule
           Eigenface &51.41	&70.61	\\
          \hline
           Fisherface &83.79	&20.62\\
          \hline
          CFA (OTF) &522.74	&32.88\\
          \hline
          CFA (OEOTF) &202.78	&30.41	\\
          \hline
          PCRC &65.27	&23.80\\
          \hline
          MS-CFB &3134.21	&82.64	\\
          \bottomrule
         \end{tabular}}
         \label{tab:TIME}
         \end{table*}

\begin{figure*}[tbh!]
         \centering
            \includegraphics[width=12 cm,height=7cm]{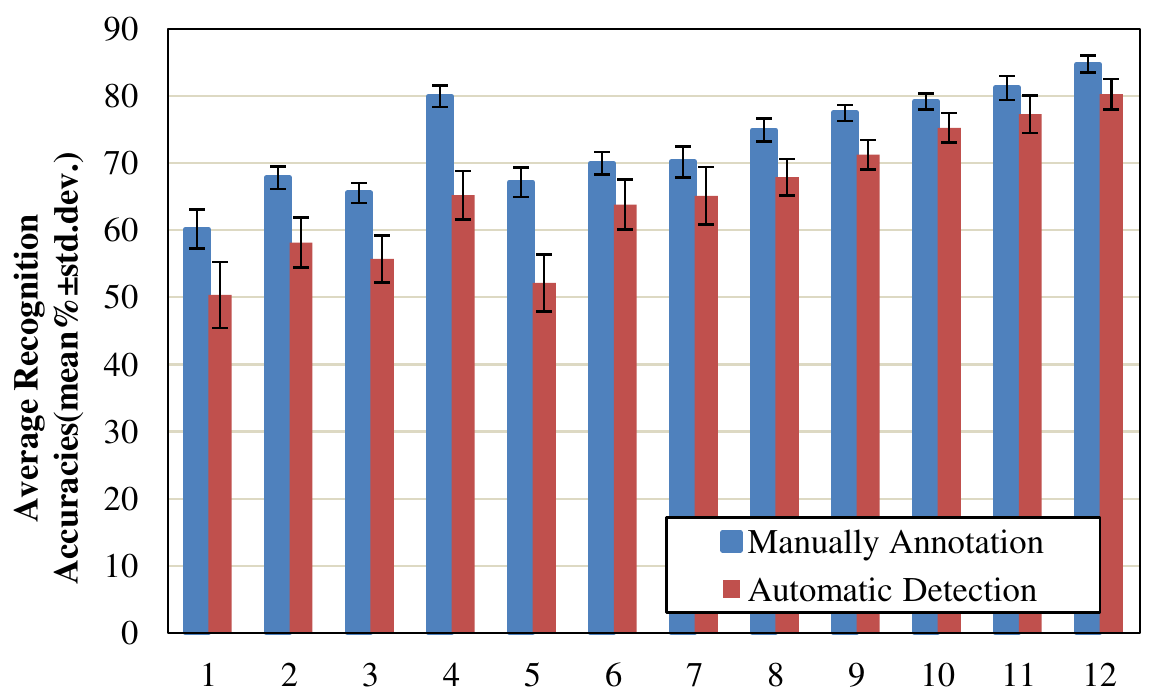}%
         \caption{ The average recognition accuracies (mean\%$\pm$std.dev.) when manual annotation and automatic detection are respectively applied. Methods 1-12 correspond to Eigenface, Fisherface, CFA(OTF), CFA (OEOTF), SRC, Block-FLD, C-LDA, HEC, BBOW, PCRC, MS-CFB (max), MS-CFB (cos), respectively.
                 }
         \label{FIG:auto}%
        \end{figure*}

As shown in Table \ref{tab:TIME}, the computational time of the proposed MS-CFB used for training is higher than that of the other algorithms. However, the computational time of MS-CFB used for recognition is comparable to that of the other algorithms (and the proposed MS-CFB achieves more accurate recognition rates when it is compared with these competing algorithms  on the CAS-PEAL R1 database).~As the training stage is usually performed offline, the computational complexity of the proposed algorithm will not constrain its applications to real-world tasks.
\\
\textcolor{red}{\subsection{Automatic Face Recognition}}
\textcolor{red}{In the above experiments, the facial part in each image is cropped and resized into the size of $80 \times 88$ based on manually annotated eye positions. However, in many real-world applications, a robust face recognition system should be a fully automatic system (it is not realistic to manually annotate the centers of eyes for each test face image). Hence, in this section, we evaluate the performance of all the competing algorithms in the applications of automatic face recognition. To be specific, we manually align and crop each face image in the training set and automatically detect, crop, and resize each image in the test set by using a popular face detector \cite{Viola2004} and an automatic eye detector \cite{Xiong2010}. A subset (includes 1,400 images of 200 individuals) of the FERET database is used for comparisons. The experimental settings used are the same as those in Section 4.4. Here, the number of training samples for each subject $t$ is set to 3.
Fig.~\ref{FIG:auto} shows the average recognition accuracies when manual annotation and automatic detection are respectively applied.}

\textcolor{red}{From Fig. \ref{FIG:auto}, we can observe that the accuracy of automatically detected positions of the centers of eyes affects the face recognition performance of all the competing algorithms. This is
due to the fact that there usually contain some spatial misalignments caused by location errors in the automatically detected face images, which leads to a negative influence on the recognition accuracy. Experimental results have verified the degrade of the recognition performance (about a $3\%\sim 10\%$ drop) with automatically detection of the centers of eyes. However,
the local-based algorithms (such as Block-FLD, BBOW, PCRC, and the proposed MS-CFB) are more robust against spatial misalignments
than the global-based algorithms (such as Eigenface, CFA, and SRC). This is because the local-based algorithms can alleviate the misalignment effects by partitioning a face image into smaller face subregions. In particular, experimental results have shown that the proposed MS-CFB gives the smallest drop on the recognition accuracy, since it effectively combines local features in an integrated framework. }

\subsection{Discussion}

From the above-mentioned experimental results, we can see that the proposed MS-CFB with the cosine similarity measure can achieve better recognition accuracies than
most competing algorithms to handle the SSS problem.
%, and MS-CFB is effective to
%extract global features based on local face subregions.
There are two reasons why MS-CFB achieves superior performance: 1) MS-CFB partitions each face image into multi-subregions and an effective learning algorithm (i.e., CFB) is applied to explore discriminative local features which are more robust to variations caused by facial expression, illumination, and pose; 2) MS-CFB extracts discriminative features in a class-specific manner, while the others extract features in a generic way.

It is worth remarking upon the performance comparisons
between different algorithms.

(1) Eigenface, which is based on PCA, extracts the most representative features
in terms of the minimal mean squared error. However, PCA is not optimal for the classification problem, which results in less effectiveness of Eigenface in face recognition. On the contrary, MS-CFB emphasizes the
correlation outputs for authentic samples while suppressing the outputs for impostor samples. Therefore, MS-CFB can extract discriminative features which effectively distinguish different classes.

(2) The projection vector obtained by Fisherface discriminates all classes. One problem of Fisherface is that it is not able to effectively discriminate two classes close to
each other since large class distances are often overemphasized (which is also known as the class separation problem \cite{Tao2009}).
In contrast, the projection vector of MS-CFB focuses on the separation between one specific class and all the other classes. As a result, MS-CFB can alleviate the class separation problem.

(3) Compared with CFA, where the correlation filter is designed in the frequency domain, the CFB used in MS-CFB only employs the feature representation in the spatial domain which improves the computational efficiency by removing the traditional Fourier transforms during the design process of a CFB. Furthermore, different from the commonly used OTF and OEOTF (which are the constrained correlation filters), the design of a CFB removes the hard constraints by using the unconstrained form so as to increase the generalization capability of the filter bank.
%4. Recent research show that sparsity is an important property for signal representation \cite{Wright2009}.
%The feature obtained by MS-CFB can be considered as a sparse feature, since only the specific CFB corresponding to the authentic class produces a peak while other CFBs produce small values. However, different from SRC, MS-CFB is designed in a totally different manner (from the feature extraction perspective). In addition, MS-CFB can be generalized to classify the unseen subjects.

(4) While most FE algorithms are required to select the optimal reduced dimension (ORD) \cite{Nie2007}, MS-CFB does not need to determine the ORD, thus improving the convenience. This is because the dimension of the feature vector obtained by MS-CFB is a fixed value (which is equal to the number of classes in a training set).
%6. One desirable property of MS-CFB is that the online updating is convenient when new classes are added.
%The previous trained correlation filter banks do not require retraining.
Moreover, compared with popular local-based FE algorithms (such as HEC and PCRC), where the local FE step and the combination of local subregions are performed as two independent processes, MS-CFB unifies these two processes in an effective framework.
%\vspace{-0.5cm}
\section{Conclusions and Future Work}
%\vspace{-0.2cm}
In this paper, we have presented an effective feature extraction algorithm called MS-CFB and applied it to the task of face recognition. MS-CFB unifies the local feature extraction step and the combination of different face subregions in an integrated framework. The key idea of MS-CFB is that, instead of extracting local features independently for each face subregion, the local feature extraction steps for different face subregions are combined to give optimal overall correlation outputs.
We have evaluated MS-CFB under different conditions, including variations in illumination, facial expression, and pose, as well as dealing with the SSPP problem. Experimental results have shown that MS-CFB outperforms most state-of-the-art feature extraction algorithms, such as SRC, HEC, and PCRC, on popular face databases for solving the SSS problem.

As mentioned in our experiments, the multi-block division strategy (based on rectangle blocks) used in the proposed algorithm cannot handle face recognition with large pose variations well due to the fact that all face images are manually aligned according to the eye positions. Recent work has demonstrated that the usage of irregular subregions can be helpful to improve face recognition performance. For instance, Kumar et al.~\cite{Kumar2011} defined 10 subregions with different shapes (e.g., rectangles, eclipses, polygons, etc.) corresponding to functional parts of a face (such as the nose, mouth, eye) in recognition. Hence, how to design adaptive face subregions to improve the performance of MS-CFB under large pose variations is an interesting direction of our future work. In addition, we are interested in extending the idea of MS-CFB to the task of facial expression recognition and other biometric recognition applications.

\section*{Acknowledgments}
The authors would like to thank the anonymous reviewers for their constructive comments. This work was supported by the National Natural Science Foundation of China under Grants 61201359 and 61170179, by the Natural Science Foundation of Fujian Province of China under Grant 2012J05126, by the Specialized Research Fund for the Doctoral Program of Higher Education of China under Grant 20110121110033.

\bibliographystyle{model1-num-names}


\newpage
\noindent AUTHOR BIOGRAPHY
\\
\textbf{Yan Yan} is currently an assistant professor at Xiamen University, China. He received the Ph.D.~degree in Information and Communication Engineering from Tsinghua University, China, in 2009. He worked at Nokia Japan R\&D center as a research engineer (2009-2010) and Panasonic Singapore Lab as a project leader (2011). His research interests include image recognition and machine learning.
\\
\\
\textbf{Hanzi Wang} is currently a Distinguished Professor and ``Minjiang Scholar" at Xiamen University, China, and an Adjunct Professor at the University of Adelaide, Australia. He has published more than 60 papers in major international journals and conferences including the IEEE T-PAMI, IJCV, ICCV, CVPR, ECCV, NIPS, MICCAI, etc. He is an Associate Editor for IEEE T-CSVT. He is a Senior Member of the IEEE and is currently the General Chair of the 6th International Conference on Internet Multimedia Computing and Service (ICIMCS 2014). His research interests are concentrated on computer vision and pattern recognition.
\\
\\
\textbf{David Suter} is currently a professor in the school of Computer Science, The University of Adelaide. He is head of the School of Computer Science. He served on the Australian Research Council (ARC) College of Experts from 2008-2010. He is on the editorial boards of IJCV. He has previously served on the editorial boards of MVA and IJIG. He was General co-Chair or the Asian Conference on Computer Vision (Melbourne 2002) and is currently co-Chair of the IEEE International Conference on Image Processing (ICIP2013). His main research interests are computer vision and pattern recognition.

\end{document}